\documentclass[10pt,twocolumn,letterpaper]{article}

\usepackage{subcaption}
\usepackage{cvpr}
\usepackage{times}
\usepackage{epsfig}
\usepackage{graphicx}
\usepackage{amsmath}
\usepackage{amssymb}

\usepackage{xspace}
\usepackage{multirow}
\usepackage{bbm}
\usepackage{afterpage}

\usepackage{multirow}
\usepackage{nopageno}

\usepackage{algorithm}
\usepackage[noend]{algpseudocode}
\usepackage{tabularx}
\usepackage{makecell}
\usepackage{microtype}
\usepackage{booktabs}
\usepackage{textcomp}
\usepackage{colortbl}
\usepackage[table]{xcolor}
\usepackage{romannum}

\usepackage[pagebackref=true,breaklinks=true,letterpaper=true,colorlinks,bookmarks=false]{hyperref}

\cvprfinalcopy %

\def\cvprPaperID{3419} %

\ifcvprfinal\fi

\newcommand{\Dataset}{{MannequinChallenge}\xspace}
\newcommand{\DatasetShort}{{MC}\xspace}

\newcommand{\sifull}{{\bf si-full}\xspace}
\newcommand{\sihuman}{{\bf si-hum}\xspace}
\newcommand{\siinter}{{\bf si-inter}\xspace}
\newcommand{\siintra}{{\bf si-intra}\xspace}
\newcommand{\sienv}{{\bf si-env}\xspace}

\newcommand{\sirmse}{\text{si-RMSE}\xspace}

\DeclareMathOperator*{\argmax}{arg\,max}

\newcommand{\Lsi}{\mathcal{L}_\mathsf{si}}
\newcommand{\Lmse}{\mathcal{L}_\mathsf{MSE}}
\newcommand{\Lgrad}{\mathcal{L}_\mathsf{grad}}

\newcommand{\Lsm}{\mathcal{L}_\mathsf{sm}}

\newcommand{\humanR}{\mathcal{H}}
\newcommand{\envR}{\mathcal{E}}

\newcommand{\Iref}{I^r}
\newcommand{\Isrc}{I^s}

\newcommand{\DPP}{D_{\text{pp}}}
\newcommand{\DMVS}{D_{\text{MVS}}}

\begin{document}

\title{Learning the Depths of Moving People by Watching Frozen People\vspace{-0.5cm}}
\author{
Zhengqi Li${}$\qquad
Tali Dekel${}$\qquad
Forrester Cole${}$ \qquad
Richard Tucker${}$ \\[1mm]
Noah Snavely${}$ \qquad
Ce Liu${}$ \qquad
William T. Freeman${}$
\\[2mm]
Google Research
\vspace{-2mm}
}
\maketitle
\begin{figure}[t]
\vspace{-2.7in}
\centering
\begin{minipage}{\textwidth}
	\centering
\includegraphics[width=1\textwidth]{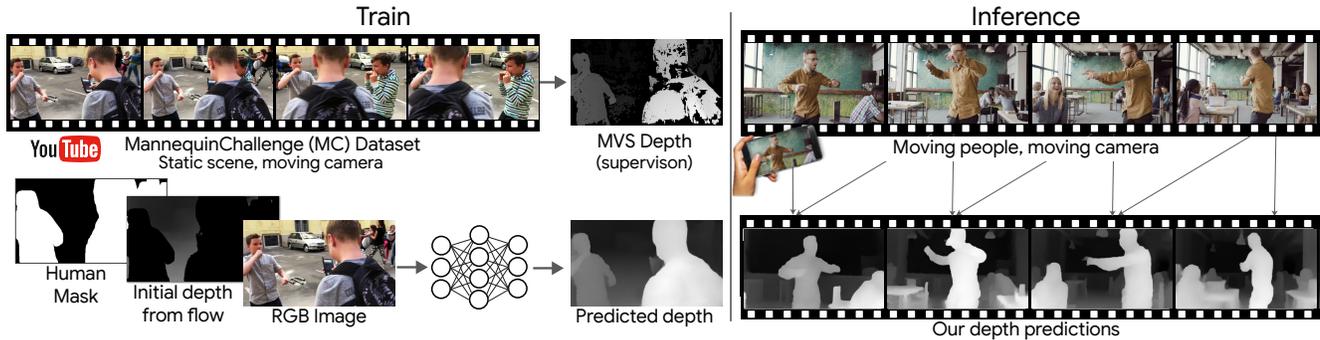}\vspace{-0.1cm}
\caption{Our model predicts dense depth when both an ordinary camera and people in the scene are freely moving (right). We train our model on our new \emph{\Dataset} dataset---a collection of Internet videos of people imitating mannequins, i.e., freezing in diverse, natural poses, while a camera tours the scene (left). Because people are \emph{stationary}, geometric constraints hold; this allows us to use multi-view stereo to estimate depth which serves as supervision during training.{\protect\footnotemark}}
\vspace{-0.4cm}
\label{fig:teaser}
\end{minipage}
\end{figure} 
\footnotetext{In all figures, we use inverse depth maps for visualization purposes, and refer to them as depth maps.}

\begin{abstract}
We present a method for predicting dense depth in scenarios where both a monocular camera and people in the scene are freely moving. Existing methods for recovering depth for dynamic, non-rigid objects from monocular video impose strong assumptions on the objects' motion and may only recover sparse depth. In this paper, we take a data-driven approach and learn human depth priors from a new source of data: thousands of Internet videos of people imitating mannequins, i.e., freezing in diverse, natural poses, while a hand-held camera tours the scene. Because people are stationary, training data can be generated using multi-view stereo reconstruction. At inference time, our method uses motion parallax cues from the static areas of the scenes to guide the depth prediction. 
We demonstrate our method on real-world sequences of complex human actions captured by a moving hand-held camera, show improvement over state-of-the-art monocular depth prediction methods, and show various 3D effects produced using our predicted depth.

\end{abstract}

\section{Introduction}

A hand-held camera viewing a dynamic scene is a common scenario in modern photography. Recovering dense geometry in this case is a challenging task: 
moving objects violate the epipolar constraint used in 3D vision, and are often treated as noise or outliers in existing structure-from-motion (SfM) and multi-view stereo (MVS) methods. %
Human depth perception, however, is not easily fooled by object motion---rather, we maintain a feasible interpretation of the objects' geometry and depth ordering even if both objects and the observer are moving, and even when the scene is observed with just one eye~\cite{howard2002seeing}. In this work, we take a step towards achieving this ability computationally. 

We focus on the task of predicting accurate, dense depth from ordinary videos where both the camera and \emph{people} in the scene are \emph{naturally moving}. We focus on humans for two reasons: i) in many applications (e.g., augmented reality), humans constitute the salient objects in the scene, and ii) human motion is articulated and difficult to model. By taking a data-driven approach, we avoid the need to explicitly impose assumptions on the shape or deformation of people, but rather learn these priors from data.

\begin{figure*}[t!]
    \centering
    \includegraphics[width=1\textwidth]{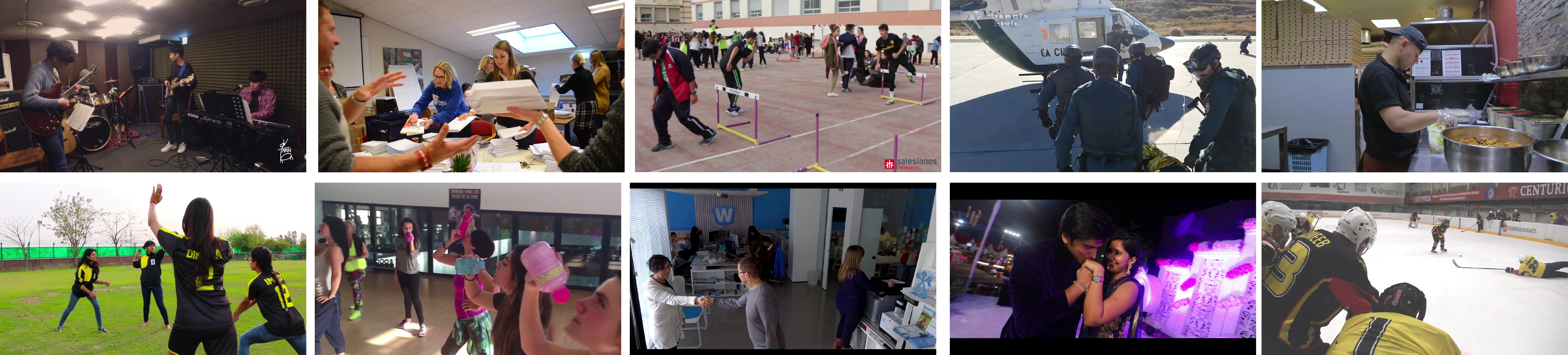}
    \caption{{\bf Sample images from Mannequin Challenge videos}. Each image is a frame from a video sequence in which the camera is moving but \emph{humans are all static}. The videos span a variety of natural scenes, poses, and configuration of people.  \vspace{-0.3cm}}
    \label{fig:dataset}
\end{figure*}

Where do we get data to train such a method? Generating high-quality synthetic data in which both the camera and the people in the scene are naturally moving is very challenging.
Depth sensors (e.g., Kinect) can provide useful data, but such data is typically limited to indoor environments and requires significant manual work in capture and process. Furthermore, it is difficult to gather people of different ages and genders with diverse poses at scale. Instead, we derive data from a surprising source: YouTube videos
in which people imitate mannequins, i.e., freeze in elaborate, natural poses, while a hand-held camera tours the scene (Fig.~\ref{fig:dataset}). These videos comprise our new \emph{\Dataset (\DatasetShort)} dataset, which we plan to release for the research community. Because the entire scene, including the people, is stationary, we estimate camera poses and depth using SfM and MVS, and use this derived 3D data as supervision for training.

In particular, we design and train a deep neural network that takes an input RGB image, a mask of human regions, and an initial depth of the environment (i.e., non-human regions), and outputs a dense depth map over the \emph{entire} image,  both the environment and the people (see Fig.~\ref{fig:teaser}). Note that the initial depth of the environment is computed using motion parallax between two frames of the video, providing the network with information not available from a single frame.  Once trained, our model can handle natural videos with arbitrary camera and human motion.

We demonstrate the applicability of our method on a variety of real-world Internet videos, shot with a hand-held camera, depicting complex human actions such as walking, running, and dancing. Our model predicts depth with higher accuracy than state-of-the-art monocular depth prediction and motion stereo methods. We further show how our  depth maps can be used to produce various 3D effects such as synthetic depth-of-field, 
depth-aware inpainting, and inserting virtual objects into the 3D scene with correct occlusion. 

In summary, our contributions are: i) a new source of data  for depth prediction consisting of a large number of Internet videos in which the camera moves around people ``frozen'' in natural poses, along with a methodology for generating accurate depth maps and camera poses; and ii) a deep-network-based model designed and trained to predict dense depth maps in the challenging case of simultaneous camera motion and complex human motion.

\section{Related Work}

\noindent \textbf{Learning-based depth prediction.}
Numerous algorithms, based on both supervised and unsupervised learning, have recently been proposed for predicting dense depth from a single RGB image~\cite{xu2018monocular, laina2016deeper, fu2018deep, eigen2014depth, chen2016single, li2018megadepth, schonberger2016pixelwise, Godard2017UnsupervisedMD, Zhou2017UnsupervisedLO, Yin2018GeoNetUL, mahjourian2018unsupervised, Wang_2018_CVPR}. %
Some recent learning based methods also consider multiple images, either assuming known camera poses~\cite{huang2018deepmvs, yao2018mvsnet} or simultaneously predicting camera poses along with depth~\cite{ummenhofer2017demon, zhou2018deeptam}. However, none of them is designed to predict the depth of dynamic objects, which is the focus of our work.

\medskip
\noindent\textbf{Depth estimation for dynamic scenes.}
RGBD data has been widely used for 3D modeling of dynamic scenes ~\cite{newcombe2015dynamicfusion, zollhofer2014real, ye2014real, Dou2016Fusion4DRP, Innmann2016VolumeDeformRV}, but only a few methods attempt to estimate depth from a monocular camera. Several  methods have been proposed to reconstruct sparse geometry of a dynamic scene \cite{Park20103DRO, Zheng2015SparseD3, Simon2017KroneckerMarkovPF, Vo2016SpatiotemporalBA}. Russell~\etal~\cite{russell2014video} and Ranftl~\etal~\cite{ranftl2016dense} suggest motion/object segmentation based algorithms to  decompose a dynamic scene into piecewise rigid parts. 
However, these methods impose strong assumptions of the object's motion that are violated by articulated human motion. Konstantinos~\etal ~\cite{rematas2018soccer} predict depth of moving soccer players using synthetic training data from FIFA video games. However, their method is limited to soccer players, and cannot handle general people in the wild.

\medskip
\noindent \textbf{RGBD data for learning depth.}
There are a number of RGBD datasets of indoor scenes, captured using  depth sensors~\cite{silberman2012indoor, chang2017matterport3d, dai2017scannet, xiao2013sun3d} or synthetically rendered~\cite{song2016ssc}. However, none of these datasets provide depth supervision for moving people in natural environments. %
Several action recognition methods use depth sensors to capture human actions~\cite{zhu2014evaluating, shrivastava2017learning, mees16iros, ni2011rgbd}, but most use a static camera and provide only a limited number of indoor scenes. REFRESH~\cite{lv2018learning} is a recent semi-synthetic scene flow dataset created by overlaying animated people on NYUv2 images. Here too, the data is limited to interiors and consists of synthetic humans placed in unrealistic configurations with their surrounding. 

\medskip
\noindent\textbf{Human shape and pose prediction.}
Recovery of a posed 3D human mesh from a single RGB image has attracted significant attention~\cite{Lassner2017UniteTP, Gler2018DensePoseDH, kanazawa2018end, Bogo2016KeepIS, Pavlakos2017CoarsetoFineVP, Mehta2017VNectR3}. Recent methods achieve impressive results on natural images spanning a variety of poses. However, such methods only model the human body, disregarding hair, clothing, and the non-human parts of the scenes. %
Finally, many of these methods rely on correctly detecting human keypoints, requiring most of the body to be within the frame.

\section{\Dataset Dataset} \label{sec:dataset}

The \emph{Mannequin Challenge}~\cite{wikimannequinchallenge} is a popular video trend in which people freeze in place---often in an interesting pose---while the camera operator moves around the scene filming them (e.g., Fig.~\ref{fig:dataset}). Thousands of such videos have been created and uploaded to YouTube since late 2016.  To the extent that people succeed in staying still during the videos, we can assume the scenes are static and obtain accurate camera poses and depth information by processing them with SfM and MVS algorithms. We found around 2,000 candidate videos for which this processing is possible.  These videos comprise our new \emph{\Dataset (\DatasetShort) Dataset}, which spans a wide range of scenes with people of different ages, naturally posing in different group configurations. We next describe in detail how we process the videos and derive our training data.

\medskip
\noindent \textbf{Estimating camera poses.}
Following a similar approach to Zhou~\etal~\cite{zhou2018stereo}, we use ORB-SLAM2~\cite{mur2017orb} to identify trackable sequences in each video and to estimate an initial camera pose for each frame. At this stage, we process a lower-resolution version of the video for efficiency, and set the field of view to 60 degrees (typical value for modern cell-phone cameras). We then reprocess each sequence at a higher resolution using a visual SfM system~\cite{schonberger2016structure}, which refines the initial camera poses and intrinsic parameters. This method extracts and matches features across frames, then performs a global bundle adjustment optimization.  Finally, sequences with non-smooth camera motion are removed using the technique of Zhou~\etal~\cite{zhou2018stereo}.

\medskip
\noindent \textbf{Computing dense depth with MVS.} 
Once the camera poses for each clip are estimated, we then
reconstruct each scene's dense geometry. In particular, we recover
per-frame dense depth maps using COLMAP, a state-of-the-art MVS system~\cite{schonberger2016pixelwise}.

Because our data consists of challenging Internet videos that involve camera motion blur, shadows, reflections, etc., the raw depth maps estimated by MVS are often too noisy for training purposes. We address this issue by a careful depth filtering mechanism. We first filter outlier depths using the depth refinement method of~\cite{li2018megadepth}. %
We further remove erroneous depth values by considering the consistency of the MVS depth and the depth obtained from motion parallax between two frames. Specifically, for each frame, we compute a normalized error $\Delta(\mathbf{p})$ for every valid pixel $\mathbf{p}$: 
 \begin{align}
    \Delta(\mathbf{p}) = \frac{|\DMVS(\mathbf{p}) - \DPP(\mathbf{p})|}{\DMVS(\mathbf{p}) + \DPP(\mathbf{p})} \label{eq:clean}
\end{align}
where $\DMVS$ is the depth map obtained by MVS and $\DPP$ is the depth map computed from two-frame motion parallax (see Sec.~\ref{sec:pp}). Depth values for which $\Delta (\mathbf{p}) > \delta$ are removed, where we empirically set $\delta = 0.2$.%

\begin{figure*}[t!]
 \centering
	\vspace*{0.1em} 
    \begin{subfigure}[b]{0.19\textwidth}
        \includegraphics[width=\textwidth]{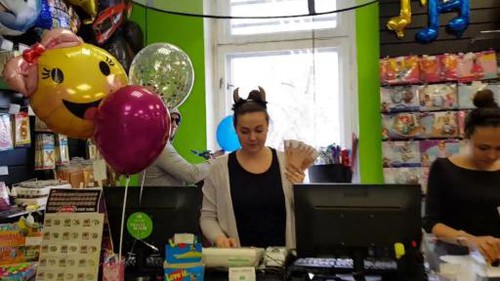}
   \end{subfigure} \hspace*{-0.7em}
   ~
    \begin{subfigure}[b]{0.19\textwidth}
       \includegraphics[width=\textwidth]{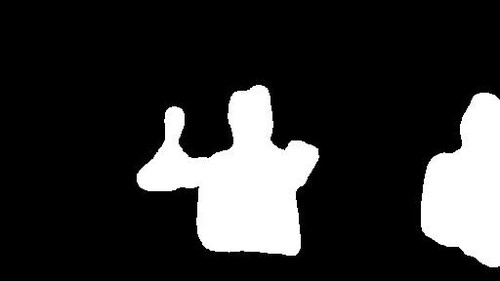}
    \end{subfigure} \hspace*{-0.7em}
    ~
    \begin{subfigure}[b]{0.19\textwidth}
        \includegraphics[width=\textwidth]{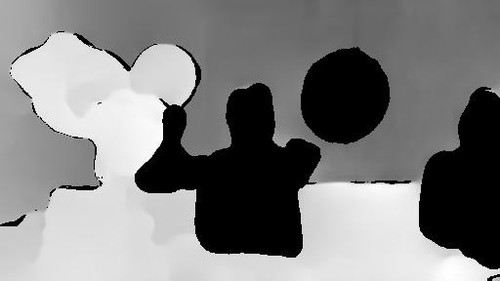}
    \end{subfigure} \hspace*{-0.7em}
    ~
	\vspace*{0.1em}
    \begin{subfigure}[b]{0.19\textwidth}
        \includegraphics[width=\textwidth]{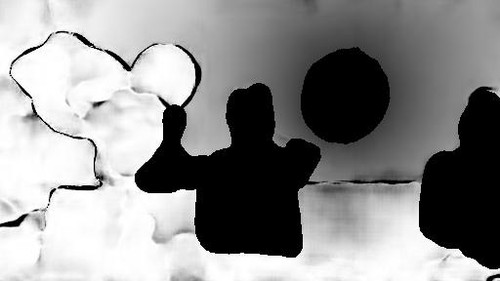}
   \end{subfigure} \hspace*{-0.7em}
    ~
    \begin{subfigure}[b]{0.19\textwidth}
       \includegraphics[width=\textwidth]{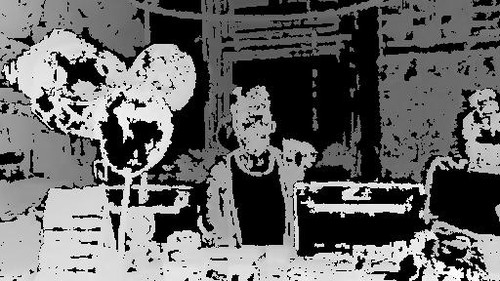}
    \end{subfigure}
    ~
    \begin{subfigure}[b]{0.19\textwidth}
        \includegraphics[width=\textwidth]{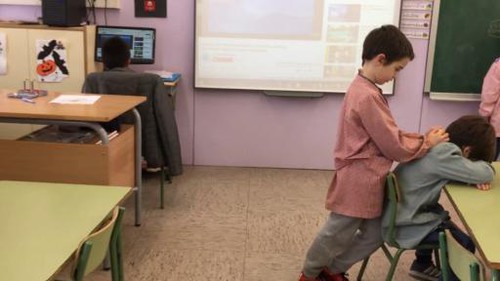}
    \end{subfigure} \hspace*{-0.7em}
    ~
	\vspace*{0.1em}
    \begin{subfigure}[b]{0.19\textwidth}
        \includegraphics[width=\textwidth]{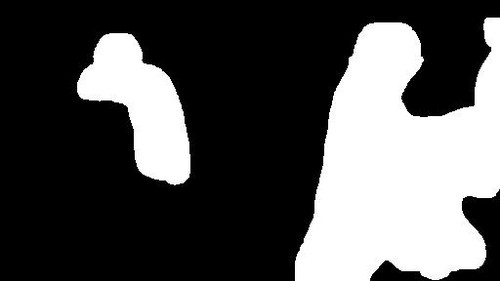}
   \end{subfigure} \hspace*{-0.7em}
    ~
    \begin{subfigure}[b]{0.19\textwidth}
       \includegraphics[width=\textwidth]{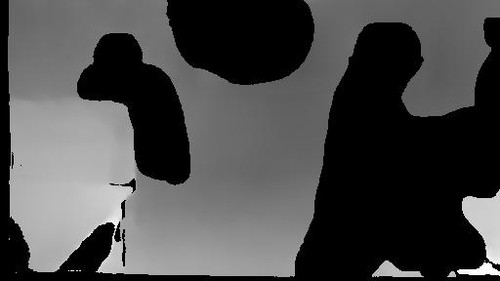}
    \end{subfigure} \hspace*{-0.7em}
    ~
    \begin{subfigure}[b]{0.19\textwidth}
        \includegraphics[width=\textwidth]{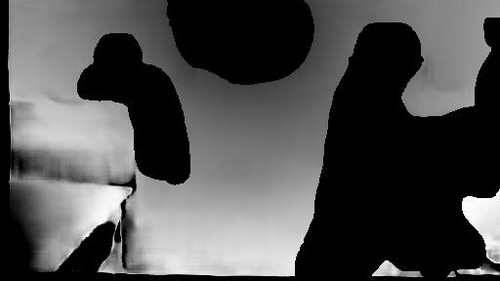}
    \end{subfigure} \hspace*{-0.7em}
    ~
    \begin{subfigure}[b]{0.19\textwidth}
        \includegraphics[width=\textwidth]{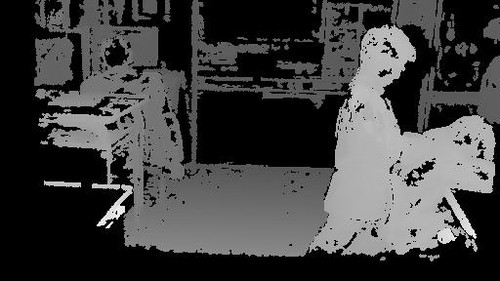}
   \end{subfigure} 
    ~
    \begin{subfigure}[b]{0.19\textwidth}
       \includegraphics[width=\textwidth]{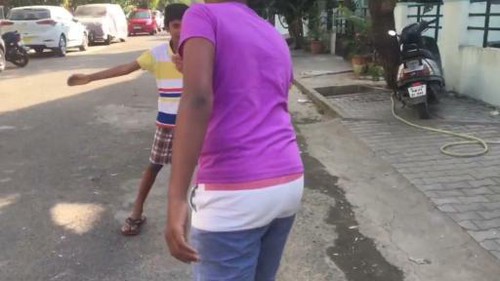}
       \caption{Reference image $\Iref$}  \vspace{-0.2em}
    \end{subfigure} \hspace*{-0.7em}
    ~
    \begin{subfigure}[b]{0.19\textwidth}
        \includegraphics[width=\textwidth]{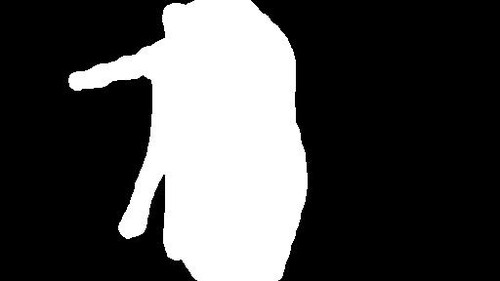}
       \caption{Human mask $M$}  \vspace{-0.2em}
    \end{subfigure} \hspace*{-0.7em}
    ~
    \begin{subfigure}[b]{0.19\textwidth}
        \includegraphics[width=\textwidth]{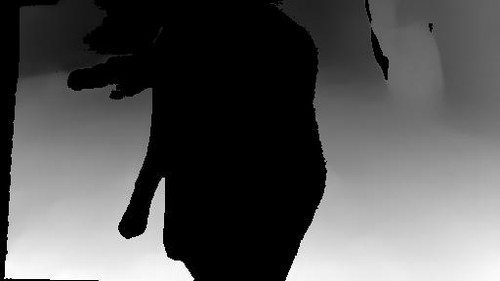}
       \caption{Input depth $D_{pp}$}  \vspace{-0.2em}
   \end{subfigure} \hspace*{-0.7em}
    ~
    \begin{subfigure}[b]{0.19\textwidth}
       \includegraphics[width=\textwidth]{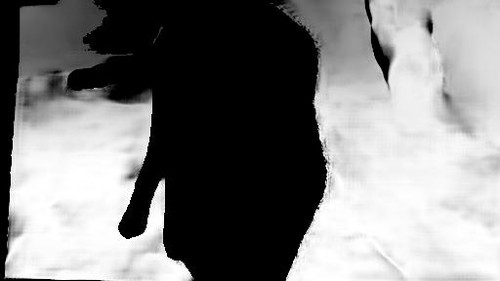}
        \caption{Input confidence $C$}  \vspace{-0.2em}
    \end{subfigure} \hspace*{-0.7em}
    ~
    \begin{subfigure}[b]{0.19\textwidth}
        \includegraphics[width=\textwidth]{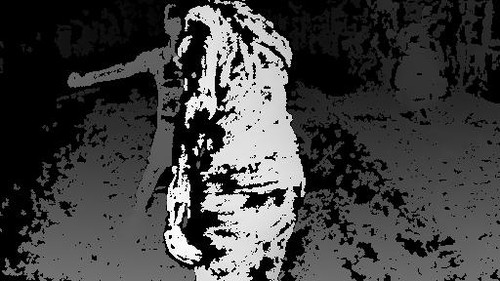}
        \caption{MVS depth $\DMVS$}  \vspace{-0.2em}
    \end{subfigure} 
    \caption{ \textbf{System inputs and training data.} The input to our network consists of: (a) RGB image,  (b) human mask,  (c) masked depth computed from motion parallax w.r.t. a selected source image, and (d) masked confidence map. Low confidence regions (dark circles) in the first two rows indicate the vicinity of the camera epipole, where depth from parallax is unreliable and is removed. The network is trained to regress to MVS depth (e).}%
    \label{fig:example_input}
    \vspace{-0.5em}%
\end{figure*}

Fig.~\ref{fig:example_input} shows sample frames from our processed sequences with corresponding estimated MVS depths after filtering. See the supplemental material for examples illustrating the effect of the proposed cleaning approach.

\medskip
\noindent \textbf{Filtering clips.} 
Several factors can make a video clip unsuitable for training. For example, people may ``unfreeze'' (start moving) at some point in the video, or the video may contain synthetic graphical elements in the background. Dynamic objects and synthetic backgrounds do not obey multi-view geometric constraints and hence are treated as outliers and filtered out by MVS, potentially leaving few valid pixels. Therefore, we remove frames where $<20\%$ of pixels have valid MVS depth after our two-pass cleaning stage.

Further, we remove frames where the estimated radial distortion coefficient $|k_1| > 0.1$ (indicative of a fisheye camera) or where the estimated focal length is $\le 0.6$ or $\ge 1.2$ (camera parameters are likely inaccurate). We keep sequences that are at least 30 frames long, have an aspect ratio of 16:9, and have a width of $\ge 1600$ pixels. Finally, we manually inspect the trajectories and point clouds of the remaining sequences and remove obviously incorrect reconstructions. Examples of removed images are shown in the supplemental material.

After processing, we obtain 4,690 sequences with a total of more then 170K valid image-depth pairs. We split our \DatasetShort dataset into training, validation and testing sets with a 80:3:17 split over clips.

\section{Depth Prediction Model} \label{sec:depth_model}

We train our depth prediction model on the \Dataset dataset in a supervised manner, i.e., by regressing to the depth generated by the MVS pipeline. 
A key question is how to structure the input to the network to allow training on frozen people but inference on freely moving people. One option is to regress from a single RGB image to depth, but this approach disregards geometric information about the static regions of the scene that is available by considering more than a single view. To benefit from such information, we input to the network a depth map for the static, non-human regions, estimated from motion parallax w.r.t. another view of the scene.

The full input to our network, illustrated in Fig.~\ref{fig:example_input}, includes a reference image  $\Iref$, a binary mask of human regions $M$, a depth map estimated from motion parallax (with human regions removed) $\DPP$, a confidence map $C$, and an optional human keypoint map $K$. We assume known, accurate camera poses from SfM during both training and inference. In an online inference setting, camera poses can be obtained by visual-inertial odometry.
Given these inputs, the network predicts a full depth map for the entire scene. To match the MVS depth values, the network must inpaint the depth in human regions, refine the depth in non-human regions from the estimated $D_{pp}$, and finally make the depth of entire scene consistent.

Our network architecture is a variant of the hourglass network of~\cite{chen2016single}, with the nearest-neighbor upsampling layers replaced by bilinear upsampling layers.

The following sections describe our model inputs and training losses in detail. In the supplemental material we provide additional implementation details and full derivations. %

\subsection{Depth from motion parallax} \label{sec:pp}
Motion parallax between two frames in a video provides our initial depth estimate for the static regions of the scene (assuming humans are dynamic while the rest of the scene is static). Given a reference image $\Iref$ and source image $\Isrc$ pair, we estimate an optical flow field from $\Iref$ to $\Isrc$ using FlowNet2.0~\cite{ilg2017flownet}. Using the relative camera poses between the two views, we compute an  initial depth map $\DPP$ from the estimated flow field, using the Plane-Plus-Parallax (P+P) representation~\cite{irani1996parallax, wulff2017optical}.

In some cases, such as forward/backward relative camera motion between the frames, the estimated depth may be ill-defined in some image regions (i.e., the epipole may be located within the image). We detect and filter out such depth values as described in Sec.~\ref{sec:confidence}.

\medskip
\noindent \textbf{Keyframe selection.} Depth from motion parallax may be ill-posed if the 2D displacement between two views is small or well-approximated by a homography (e.g., in the case of pure camera rotation). To avoid such cases, we apply a baseline criterion when selecting a reference frame $\Iref$ and a corresponding source keyframe $\Isrc$. We want the two views to have significant overlap, while having sufficient baseline. %
Formally, for each $I^r$, we find the index $s$ of $I^s$ as 
\begin{align}
    s = \argmax_{j} d^{rj} o^{rj}
\end{align}
where $d^{rj}$ is the $L_2$ distance between the camera centers of $I^r$ and its neighbor frame $I^j$. The term $o^{rj}$ is the fraction of co-visible SfM features in $I^r$ and $I^j$:
\begin{align}
  o^{rj} = \frac{2|V^r \bigcap V^j|}{|V^r| + |V^j|},
\end{align}
where $V^j$ is the set of features visible in  $I^j$. We discard pairs of frames for which $o^{rj} < \tau_o$, \ie, the fraction of co-visible features should be larger than a threshold $\tau_o$ (we set $\tau_o=0.6$), and limit the maximum frame interval to 10. We found these view selection criteria to work well in our experiments.

\begin{figure*}[t!]
 \centering
\centering
	\vspace*{0.1em} 
    \begin{subfigure}[b]{0.19\textwidth}
        \includegraphics[width=\textwidth]{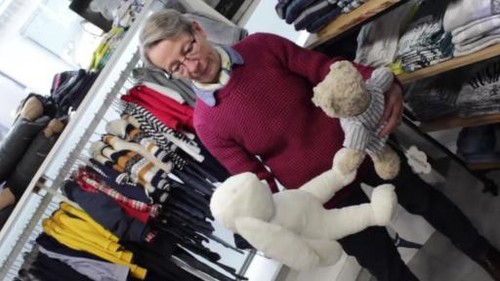}
   \end{subfigure} \hspace*{-0.7em}
    ~
    \begin{subfigure}[b]{0.19\textwidth}
        \includegraphics[width=\textwidth]{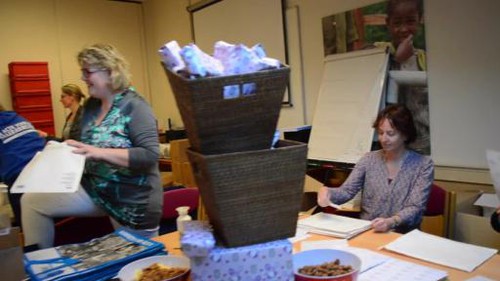}
    \end{subfigure} \hspace*{-0.7em}
    ~
    \begin{subfigure}[b]{0.19\textwidth}
       \includegraphics[width=\textwidth]{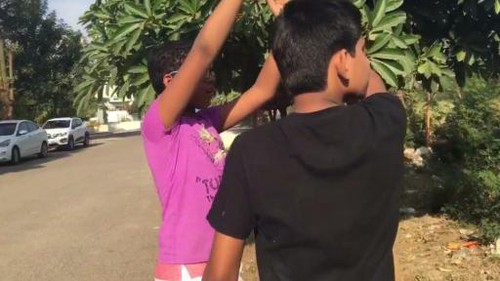}
    \end{subfigure} \hspace*{-0.7em}
    ~
    \begin{subfigure}[b]{0.19\textwidth}
        \includegraphics[width=\textwidth]{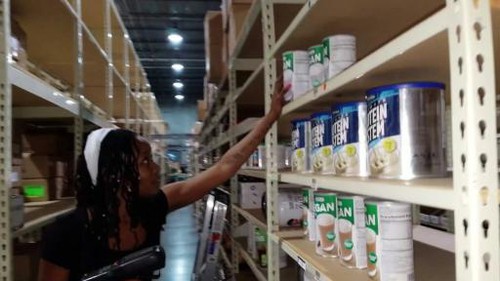}
    \end{subfigure} \hspace*{-0.7em}
    ~    
    \begin{subfigure}[b]{0.19\textwidth}
        \includegraphics[width=\textwidth]{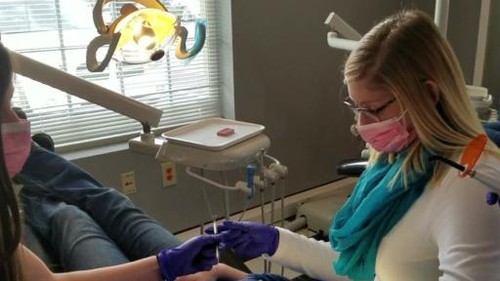}
    \end{subfigure}  
    \vspace*{0.1em} 
    \begin{subfigure}[b]{0.19\textwidth}
        \includegraphics[width=\textwidth]{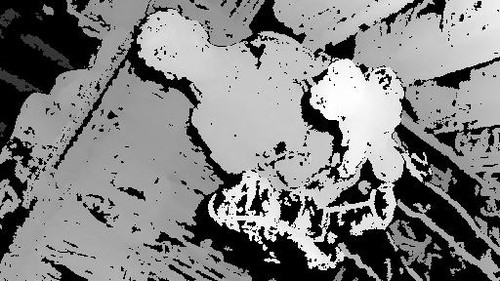}
   \end{subfigure} \hspace*{-0.7em}
    ~
    \begin{subfigure}[b]{0.19\textwidth}
        \includegraphics[width=\textwidth]{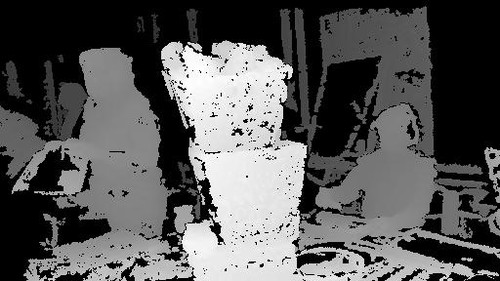}
    \end{subfigure} \hspace*{-0.7em}
    ~
    \begin{subfigure}[b]{0.19\textwidth}
       \includegraphics[width=\textwidth]{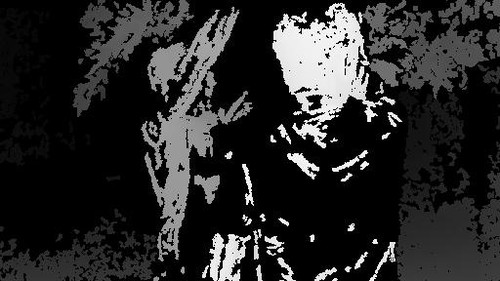}
    \end{subfigure} \hspace*{-0.7em}
    ~
    \begin{subfigure}[b]{0.19\textwidth}
        \includegraphics[width=\textwidth]{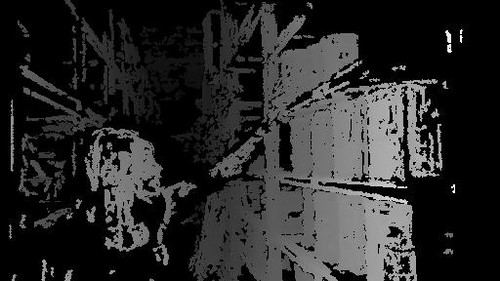}
    \end{subfigure} \hspace*{-0.7em}
    ~    
    \begin{subfigure}[b]{0.19\textwidth}
        \includegraphics[width=\textwidth]{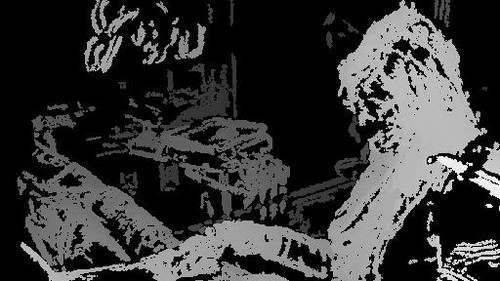}
    \end{subfigure}  
    \vspace*{0.1em} 
    \begin{subfigure}[b]{0.19\textwidth}
        \includegraphics[width=\textwidth]{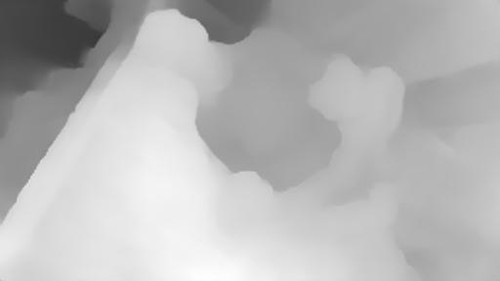}
   \end{subfigure} \hspace*{-0.7em}
    ~
    \begin{subfigure}[b]{0.19\textwidth}
        \includegraphics[width=\textwidth]{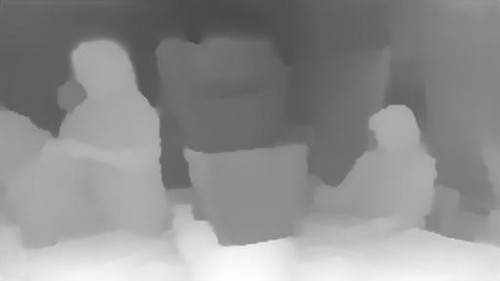}
    \end{subfigure} \hspace*{-0.7em}
    ~
    \begin{subfigure}[b]{0.19\textwidth}
       \includegraphics[width=\textwidth]{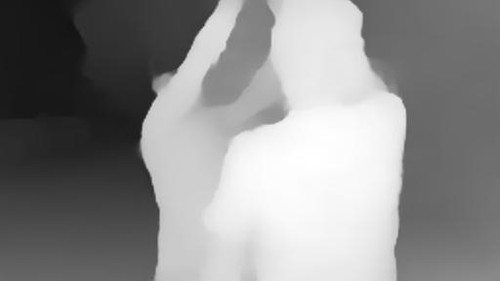}
    \end{subfigure} \hspace*{-0.7em}
    ~
    \begin{subfigure}[b]{0.19\textwidth}
        \includegraphics[width=\textwidth]{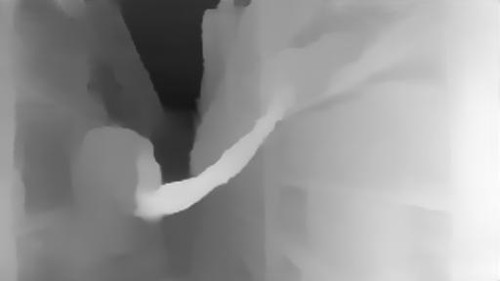}
    \end{subfigure} \hspace*{-0.7em}
    ~    
    \begin{subfigure}[b]{0.19\textwidth}
        \includegraphics[width=\textwidth]{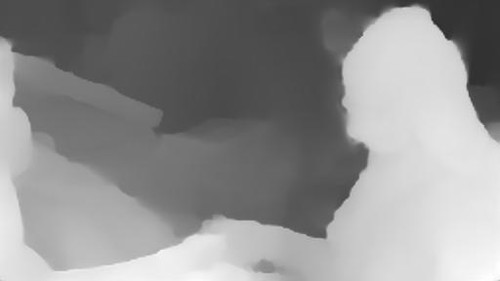}
    \end{subfigure}  
    \vspace*{0.1em} 
    \begin{subfigure}[b]{0.19\textwidth}
        \includegraphics[width=\textwidth]{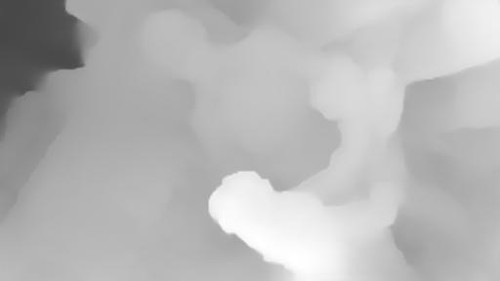}
   \end{subfigure} \hspace*{-0.7em}
    ~
    \begin{subfigure}[b]{0.19\textwidth}
        \includegraphics[width=\textwidth]{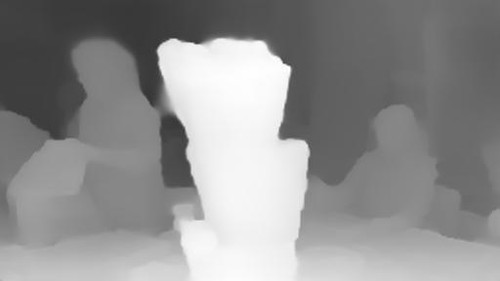}
    \end{subfigure} \hspace*{-0.7em}
    ~
    \begin{subfigure}[b]{0.19\textwidth}
       \includegraphics[width=\textwidth]{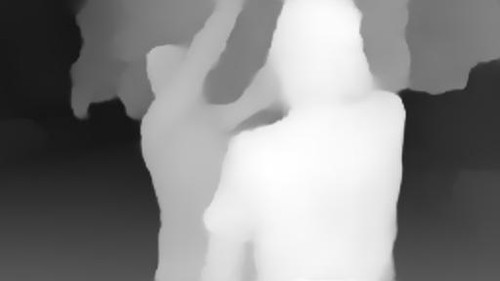}
    \end{subfigure} \hspace*{-0.7em}
    ~
    \begin{subfigure}[b]{0.19\textwidth}
        \includegraphics[width=\textwidth]{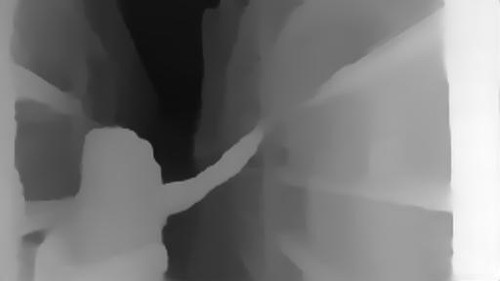}
    \end{subfigure} \hspace*{-0.7em}
    ~    
    \begin{subfigure}[b]{0.19\textwidth}
        \includegraphics[width=\textwidth]{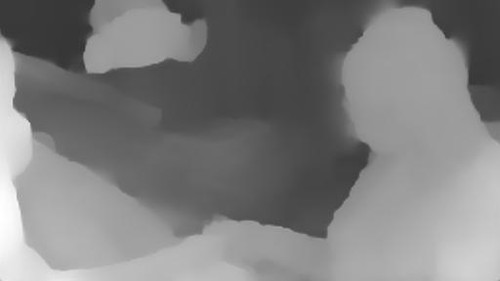}
    \end{subfigure}  
  	\caption{\textbf{Qualitative results on the \DatasetShort test set.} From top to bottom: reference images and their corresponding MVS depth (pseudo ground truth); our depth predictions using: our single view model (third row) and our two-frame model (forth row). The additional network inputs give improved performance in both human and non-human regions.}\label{fig:prediction_mc_test} %
        \vspace{-0.1in}%
\end{figure*}

\subsection{Confidence} \label{sec:confidence}
Our data consists of challenging Internet video clips with camera motion blur, shadows, low lighting, and reflections. In such cases, optical flow is often noisy~\cite{XianMonocularRD}, compounding uncertainty in the input depth map, $D_{pp}$.  We thus estimate, and input to the network, a  confidence map, $C$. This allows the network to rely more on the input depth in high-confidence regions, and potentially use it to improve its prediction in low-confidence regions. The confidence value at each pixel $\mathbf{p}$ in the non-human regions is defined as:
\begin{align}
    C(\mathbf{p}) = C_{lr}(\mathbf{p}) C_{ep}(\mathbf{p}) C_{pa}(\mathbf{p}).
\end{align}
The term $C_{lr}$ measures ``left-right'' consistency between the forward and backward flow fields. That is, $C_{lr}(\mathbf{p}) = \max \left( 0, 1 - r(\mathbf{p})^2 \right)$, where $r(\mathbf{p})$ is the forward-backward warping error. For perfectly consistent  forward and backward flows $C_{lr}\!=\!1$, while  $C_{lr}\!=\!0$ when the error is greater than $1$px.  

The term $C_{ep}$ measures how well the flow field complies with the epipolar constraint between the views~\cite{hartley2003multiple}. Specifically, $C_{ep}(\mathbf{p}) = \max \left( 0, 1 - (\gamma(\mathbf{p})/\bar{\gamma})^2 \right)$, where $\gamma(\mathbf{p})$ is the distance between the warped pixel position of $\mathbf{p}$ based on its optical flow and its corresponding epipolar line; $\bar{\gamma}$ controls the epipolar distance tolerance (we set $\bar{\gamma}=2$px in our experiments).

Finally, $C_{pa}$ assigns low confidence to pixels for which the parallax between the views is small~\cite{schonberger2016pixelwise}. This is measured by the angle $\beta(\mathbf{p})$ between the camera rays meeting at the pixel $\mathbf{p}$.  That is,  $C_{pa}(\mathbf{p})= 1- \left( \frac{\min(\bar{\beta}, \beta(\mathbf{p}) ) - \bar{\beta}}{\bar{\beta}} \right)^2$, %
where $\bar{\beta}$ is the angle tolerance (we use $\bar{\beta}=1$\textdegree\  in our experiments).

Fig.~\ref{fig:example_input}(d) shows examples of computed confidence maps. Note that human regions as well as regions for which the confidence $C(\mathbf{p}) < 0.25$  are masked out.

\subsection{Losses} 
\label{sec:losses}
We train our network to regress to depth maps computed by our data pipeline. Because the computed depth values have arbitrary scale, we use a scale-invariant depth regression loss. That is, our loss is computed on log-space depth values and consists of three terms:
\begin{align}
  \Lsi = \Lmse + \alpha_1 \Lgrad + \alpha_2 \Lsm. \label{eq:Loss} \vspace{-0.2cm}
\end{align}
\vspace{-0.8cm}\paragraph{Scale-invariant MSE.} $\Lmse $ denotes the scale-invariant mean square error (MSE)~\cite{eigen2014depth}. This term computes the squared, log-space difference in depth between two pixels in the prediction and the same two pixels in the ground-truth,  averaged over all pairs of valid pixels. 
 Intuitively, we look at all pairs of points, and penalize the difference in their  \emph{ratio}  of depth values w.r.t. ground truth. 

\vspace{-0.3cm}\paragraph{Multi-scale gradient term.}
We use a multi-scale gradient term, $\Lgrad$, which is the $L_1$ difference between the predicted log depth derivatives (in $x$ and $y$ directions) and the ground truth log depth derivatives, at multiple scales~\cite{li2018megadepth}. This term allows the network to recover sharp depth discontinuities and smooth gradient changes in the predicted depth images.
\vspace{-0.3cm}\paragraph{Multi-scale, edge-aware smoothness terms.}
To encourage smooth interpolation of depth in texture-less regions where MVS fails to recover depth, we use a simple smoothness term, $\Lsm$, which penalizes $L_1$ norm of log depth derivatives based on the first- and second-order derivatives of images and is applied at multiple scales~\cite{Wang_2018_CVPR}. This term encourages piecewise smoothness in depth regions where there is no image intensity change. %

\begin{figure*}[t!]
 \centering
\centering
    \begin{subfigure}[b]{0.135\textwidth}
        \includegraphics[width=\textwidth]{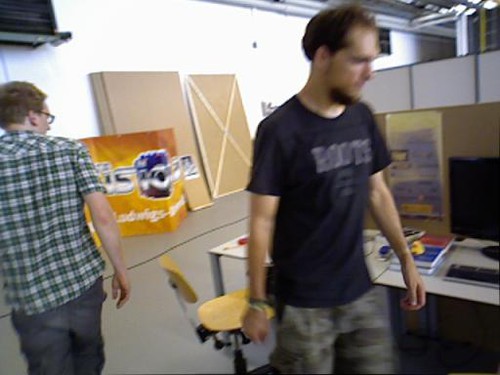}
   \end{subfigure} \hspace*{-0.8em}
    ~
    \begin{subfigure}[b]{0.135\textwidth}
        \includegraphics[width=\textwidth]{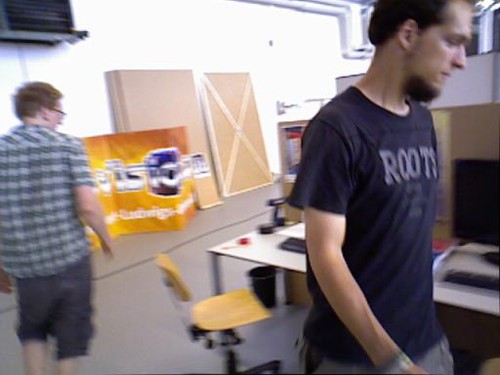}
    \end{subfigure} \hspace*{-0.8em}
    ~
    \begin{subfigure}[b]{0.135\textwidth}
       \includegraphics[width=\textwidth]{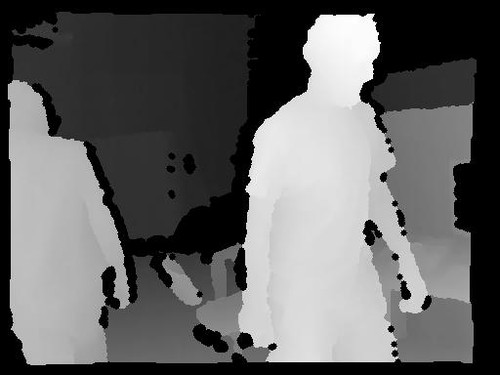}
    \end{subfigure} \hspace*{-0.8em}
    ~
    \begin{subfigure}[b]{0.135\textwidth}
        \includegraphics[width=\textwidth]{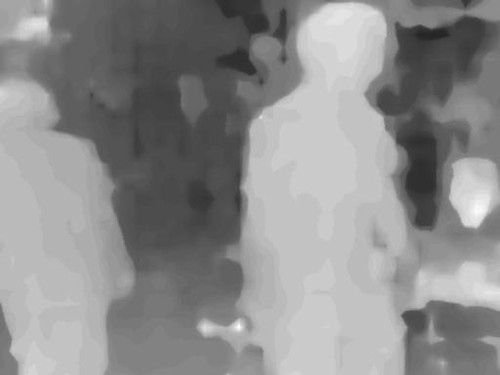}
    \end{subfigure} \hspace*{-0.8em}
    ~    
    \begin{subfigure}[b]{0.135\textwidth}
        \includegraphics[width=\textwidth]{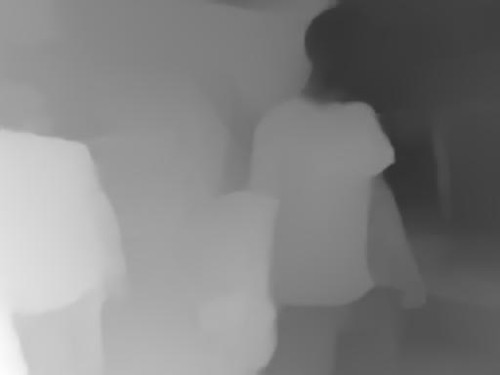}
    \end{subfigure} \hspace*{-0.8em}
    ~    
    \begin{subfigure}[b]{0.135\textwidth}
        \includegraphics[width=\textwidth]{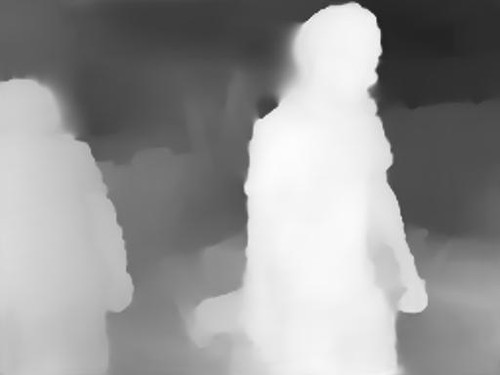}
    \end{subfigure} \hspace*{-0.8em}
    ~    
    \begin{subfigure}[b]{0.135\textwidth}
        \includegraphics[width=\textwidth]{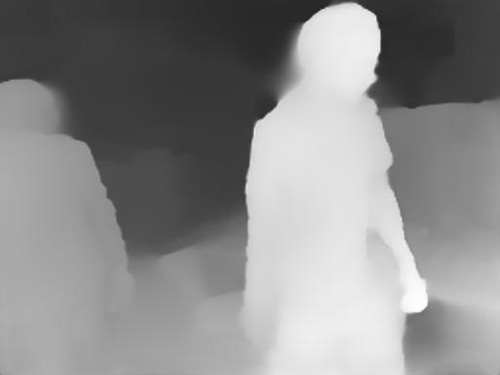}
    \end{subfigure}
    \begin{subfigure}[b]{0.135\textwidth}
        \includegraphics[width=\textwidth]{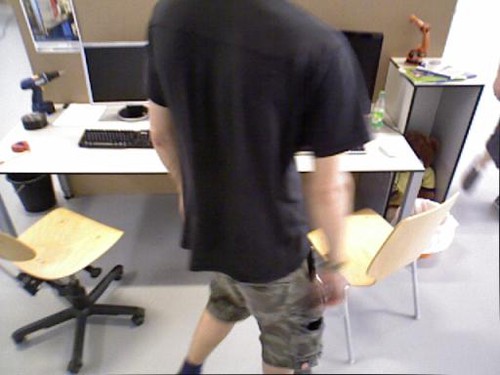}
   \end{subfigure} \hspace*{-0.8em}
    ~
    \begin{subfigure}[b]{0.135\textwidth}
        \includegraphics[width=\textwidth]{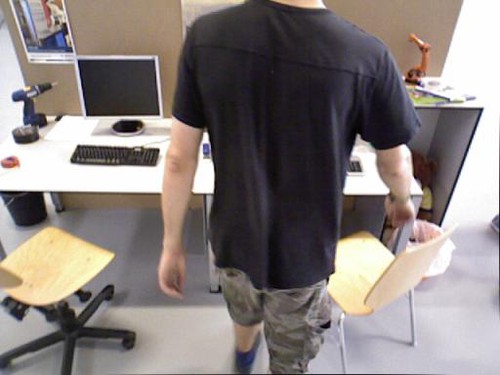}
    \end{subfigure} \hspace*{-0.8em}
    ~
    \begin{subfigure}[b]{0.135\textwidth}
       \includegraphics[width=\textwidth]{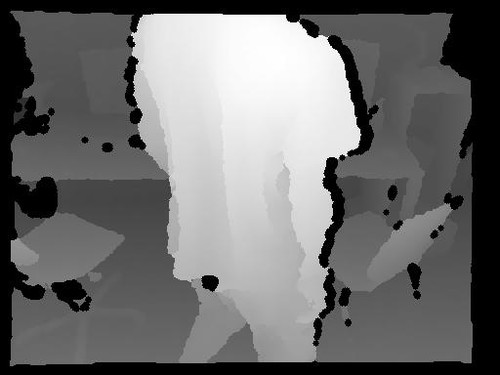}
    \end{subfigure} \hspace*{-0.8em}
    ~
    \begin{subfigure}[b]{0.135\textwidth}
        \includegraphics[width=\textwidth]{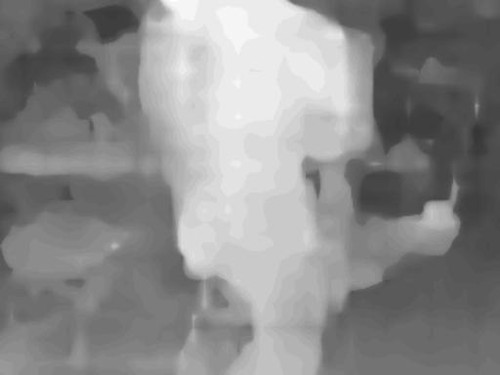}
    \end{subfigure} \hspace*{-0.8em}
    ~    
    \begin{subfigure}[b]{0.135\textwidth}
        \includegraphics[width=\textwidth]{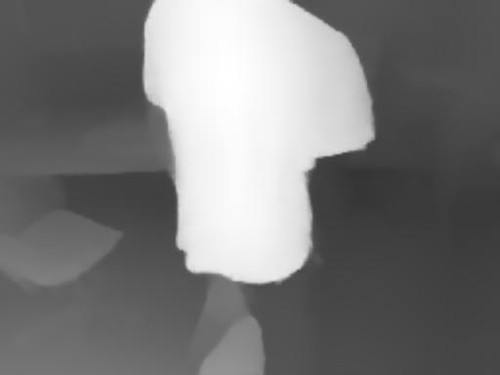}
    \end{subfigure} \hspace*{-0.8em}
    ~    
    \begin{subfigure}[b]{0.135\textwidth}
        \includegraphics[width=\textwidth]{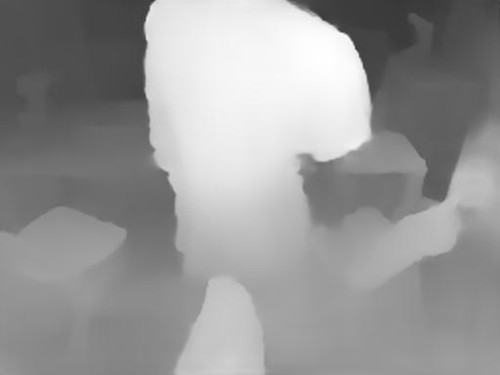}
    \end{subfigure} \hspace*{-0.8em}
    ~    
    \begin{subfigure}[b]{0.135\textwidth}
        \includegraphics[width=\textwidth]{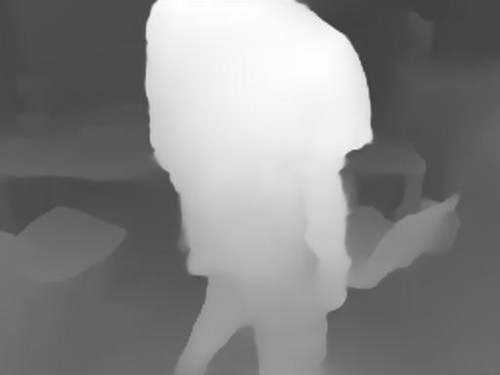}
    \end{subfigure}
    \begin{subfigure}[b]{0.135\textwidth}
        \includegraphics[width=\textwidth]{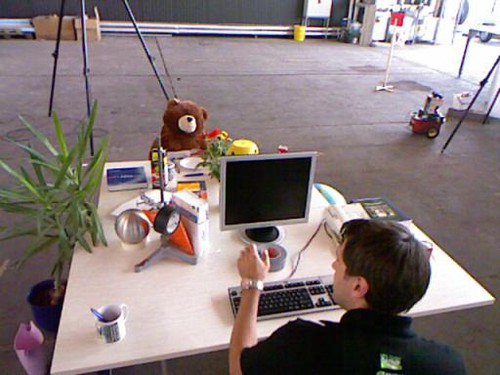}
   \end{subfigure} \hspace*{-0.8em}
    ~
    \begin{subfigure}[b]{0.135\textwidth}
        \includegraphics[width=\textwidth]{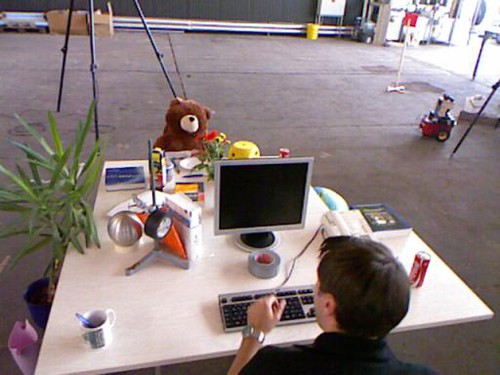}
    \end{subfigure} \hspace*{-0.8em}
    ~
    \begin{subfigure}[b]{0.135\textwidth}
       \includegraphics[width=\textwidth]{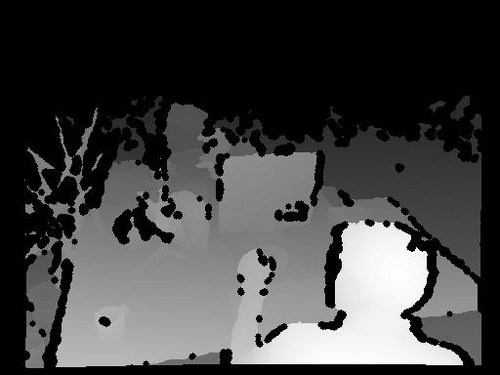}
    \end{subfigure} \hspace*{-0.8em}
    ~
    \begin{subfigure}[b]{0.135\textwidth}
        \includegraphics[width=\textwidth]{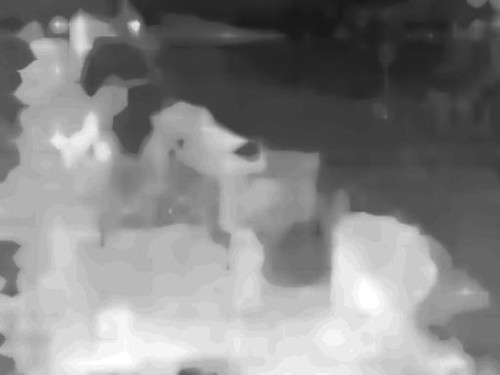}
    \end{subfigure} \hspace*{-0.8em}
    ~    
    \begin{subfigure}[b]{0.135\textwidth}
        \includegraphics[width=\textwidth]{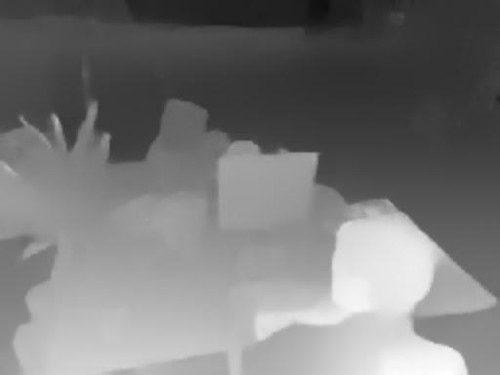}
    \end{subfigure} \hspace*{-0.8em}
    ~    
    \begin{subfigure}[b]{0.135\textwidth}
        \includegraphics[width=\textwidth]{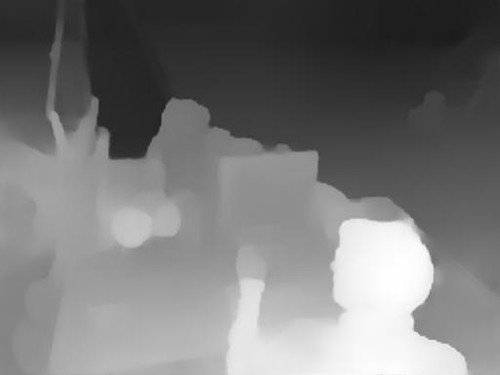}
    \end{subfigure} \hspace*{-0.8em}
    ~    
    \begin{subfigure}[b]{0.135\textwidth}
        \includegraphics[width=\textwidth]{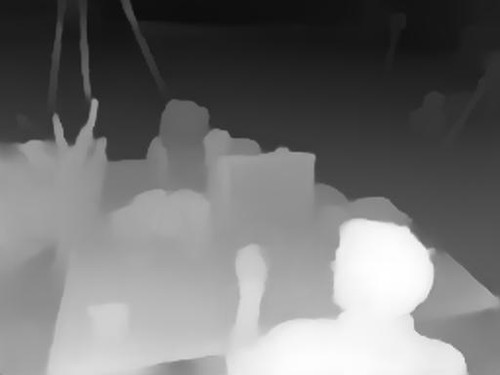}
    \end{subfigure}
    \begin{subfigure}[b]{0.135\textwidth}
        \includegraphics[width=\textwidth]{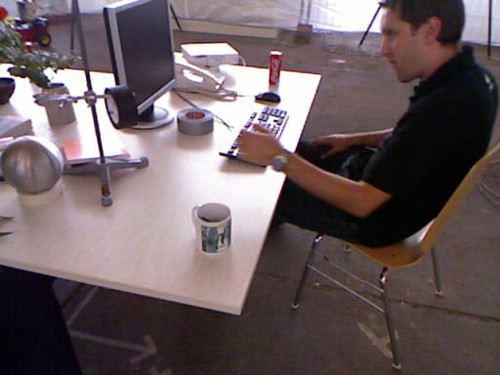}
   \end{subfigure} \hspace*{-0.8em}
    ~
    \begin{subfigure}[b]{0.135\textwidth}
        \includegraphics[width=\textwidth]{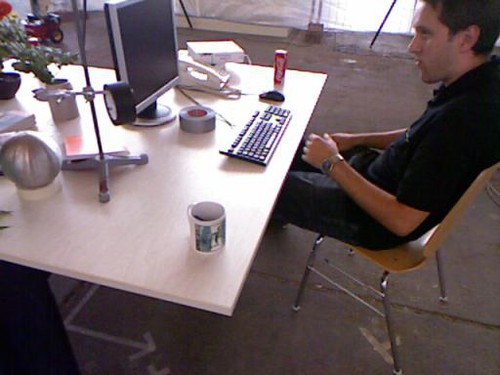}
    \end{subfigure} \hspace*{-0.8em}
    ~
    \begin{subfigure}[b]{0.135\textwidth}
       \includegraphics[width=\textwidth]{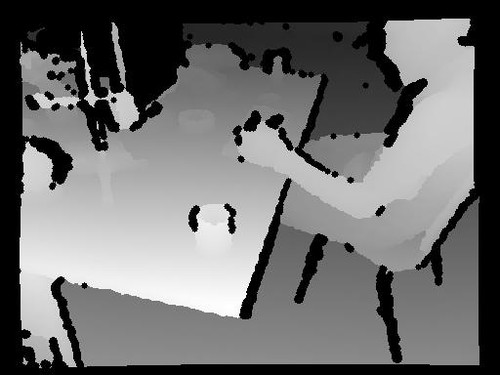}
    \end{subfigure} \hspace*{-0.8em}
    ~
    \begin{subfigure}[b]{0.135\textwidth}
        \includegraphics[width=\textwidth]{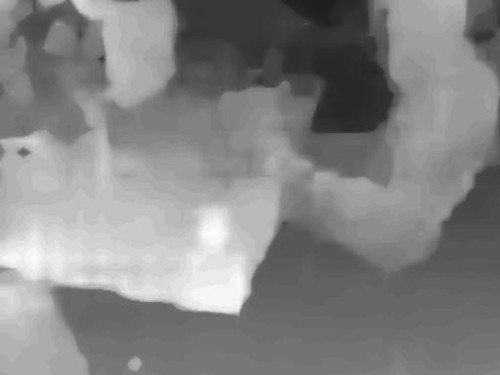}
    \end{subfigure} \hspace*{-0.8em}
    ~    
    \begin{subfigure}[b]{0.135\textwidth}
        \includegraphics[width=\textwidth]{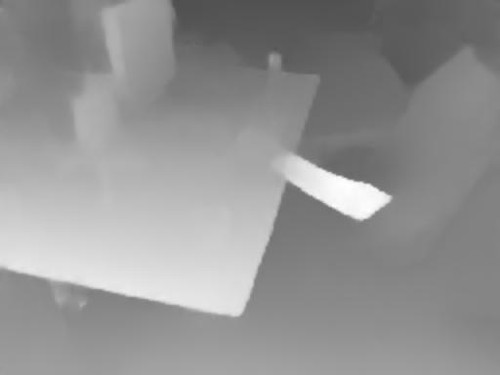}
    \end{subfigure} \hspace*{-0.8em}
    ~    
    \begin{subfigure}[b]{0.135\textwidth}
        \includegraphics[width=\textwidth]{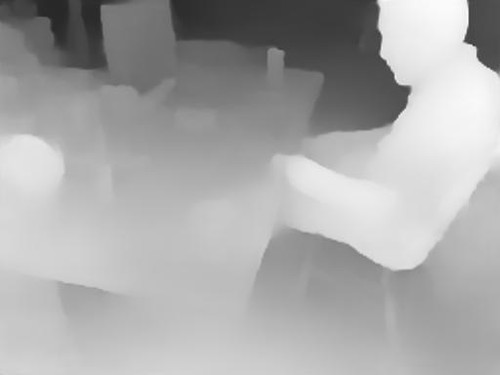}
    \end{subfigure} \hspace*{-0.8em}
    ~    
    \begin{subfigure}[b]{0.135\textwidth}
        \includegraphics[width=\textwidth]{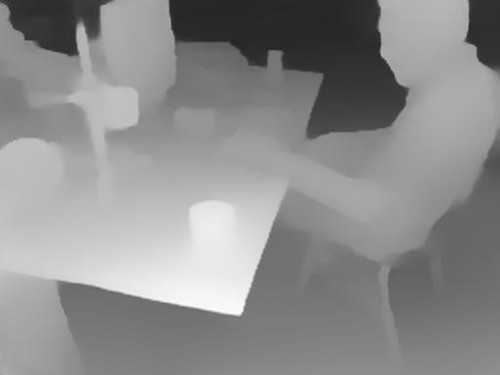}
    \end{subfigure}
    \begin{subfigure}[b]{0.135\textwidth}
        \includegraphics[width=\textwidth]{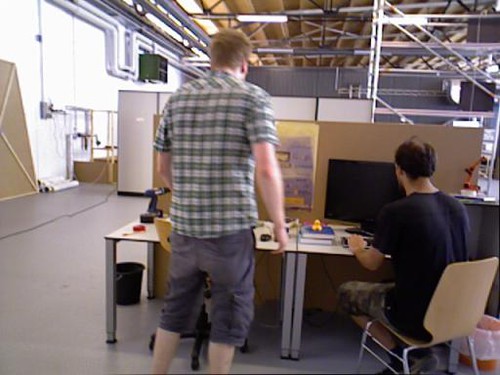}
         \caption{$\Iref$}  \vspace{-0.1em}
   \end{subfigure} \hspace*{-0.8em}
    ~
    \begin{subfigure}[b]{0.135\textwidth}
        \includegraphics[width=\textwidth]{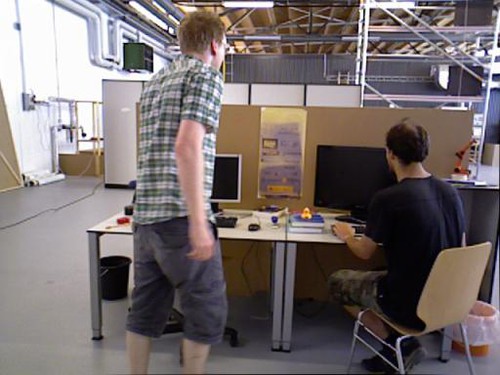}
        \caption{$\Isrc$}  \vspace{-0.1em}
    \end{subfigure} \hspace*{-0.8em}
    ~
    \begin{subfigure}[b]{0.135\textwidth}
       \includegraphics[width=\textwidth]{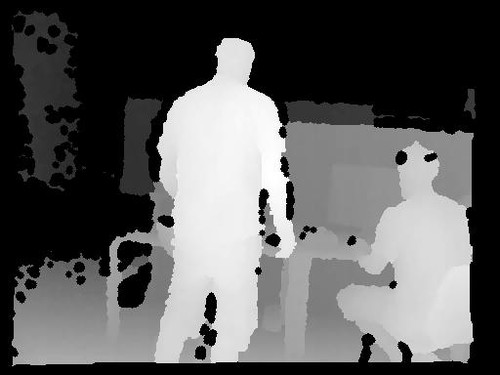}
       \caption{GT}  \vspace{-0.1em}
    \end{subfigure} \hspace*{-0.8em}
    ~
    \begin{subfigure}[b]{0.135\textwidth}
        \includegraphics[width=\textwidth]{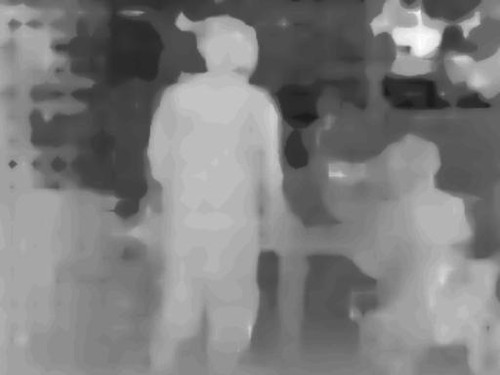}
         \caption{DORN~\cite{fu2018deep}}  \vspace{-0.1em}
    \end{subfigure} \hspace*{-0.8em}
    ~    
    \begin{subfigure}[b]{0.135\textwidth}
        \includegraphics[width=\textwidth]{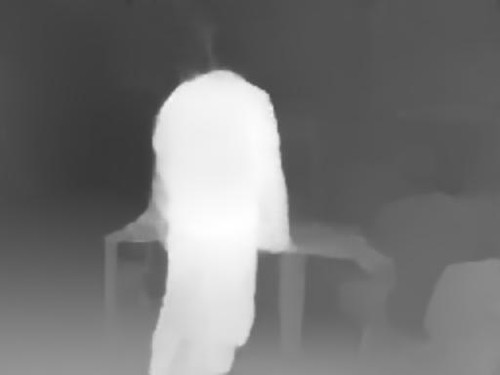}
         \caption{DeMoN~\cite{ummenhofer2017demon}} \vspace{-0.1em}
    \end{subfigure} \hspace*{-0.8em}
    ~    
    \begin{subfigure}[b]{0.135\textwidth}
        \includegraphics[width=\textwidth]{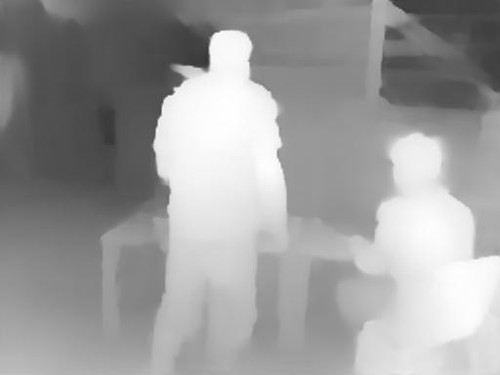}\caption{Ours (RGB)}  \vspace{-0.1em}
    \end{subfigure} \hspace*{-0.8em}
    ~    
    \begin{subfigure}[b]{0.135\textwidth}
        \includegraphics[width=\textwidth]{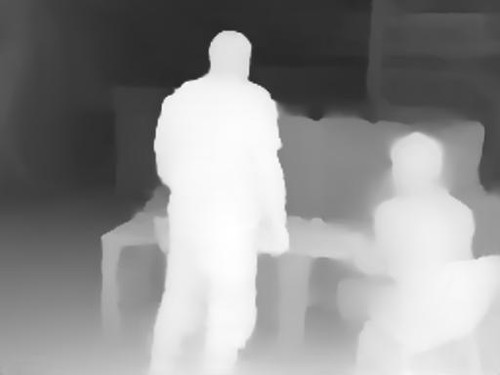} \caption{Ours (full)}  \vspace{-0.1em}
    \end{subfigure}  
  	\caption{\textbf{Qualitative comparisons on the TUM RGBD dataset.} (a) Reference images, (b) source images (used to compute our initial depth input), (c) ground truth sensor depth, (d)  single view depth prediction method DORN~\cite{fu2018deep}, (e) two-frame motion stereo DeMoN~\cite{ummenhofer2017demon}, (f-g) depth predictions from our single view and two-frame models, respectively.
  	\label{fig:tum_visual_compare}}
        \vspace{-0.15in}%
\end{figure*}

\begin{table}[t]
\centering
{\small
\begin{tabular}{lllllll}
\toprule
&\hspace{-0.2cm}Net inputs & \hspace{-0.2cm}\sifull & \hspace{-0.2cm}\sienv&\hspace{-0.2cm}\sihuman & \hspace{-0.2cm}\siintra & \hspace{-0.2cm}\siinter
\\
\midrule
\Romannum{1}.&\hspace{-0.2cm}$I$ &\hspace{-0.2cm}0.333 &\hspace{-0.2cm}0.338 & \hspace{-0.2cm}0.317 & \hspace{-0.2cm}0.264 & \hspace{-0.2cm}0.384 \\
\Romannum{2}.&\hspace{-0.2cm}$IFCM$  & \hspace{-0.2cm}0.330 &\hspace{-0.2cm}0.349 & \hspace{-0.2cm}0.312 & \hspace{-0.2cm}0.260 & \hspace{-0.2cm}0.381 \\
\Romannum{3}.\hspace{-0.5cm}&\hspace{-0.2cm}$I\DPP M$&\hspace{-0.2cm}0.255 &\hspace{-0.2cm}0.229 & \hspace{-0.2cm}0.264  & \hspace{-0.2cm}0.243 & \hspace{-0.2cm}0.285 \\
\Romannum{4}.\hspace{-0.5cm}&\hspace{-0.2cm}$I\DPP CM$  & \hspace{-0.2cm}0.232 &\hspace{-0.2cm}\textbf{0.188} & \hspace{-0.2cm}0.237 & \hspace{-0.2cm}0.221 & \hspace{-0.2cm}0.268 \\
\Romannum{5}.\hspace{-0.5cm}&\hspace{-0.2cm}$I\DPP CMK$ &\hspace{-0.2cm}\textbf{0.227} &\hspace{-0.2cm}\textbf{0.189} & \hspace{-0.2cm}\textbf{0.230} & \hspace{-0.2cm}\textbf{0.212} & \hspace{-0.2cm}\textbf{0.263} \\
\bottomrule 
\end{tabular} 
}
\vspace{0.3em}
\caption{{\bf Quantitative comparisons on the \DatasetShort test set.} Different input configurations of our model: (\Romannum{1}.) single image; (\Romannum{2}.) optical flow masked in the human region ($F$), confidence and human mask; (\Romannum{3}.) masked input depth, human mask, and additional confidence for \Romannum{4}.; in \Romannum{5}, we also input human keypoints. Lower is better for all metrics.}
\label{tb:mc_test}
\vspace{-1.3em}
\end{table}

\section{Results}
We tested our method quantitatively and qualitatively and compare it
with several state-of-the-art single-view and motion-based
depth prediction algorithms. We show additional qualitative
results on challenging Internet videos with complex human
motion and natural camera motion, and demonstrate how our
predicted depth maps can be used for several visual effects. %

\smallskip
\noindent{\bf Error metrics.} 
We measure error using the scale-invariant RMSE ($\sirmse$), equivalent to $\sqrt{\Lmse}$, described in Sec.~\ref{sec:losses}.
We evaluate $\sirmse$ on 5 different regions: \sifull measures the error between all pairs of pixels, giving the overall accuracy across the entire image; \sienv measures pairs of pixels in non-human regions $\envR$, providing depth accuracy of the environment; and \sihuman measures pairs where at least one pixel lies in the human region $\humanR$, providing depth accuracy for people. \sihuman can further be divided into two error measures: \siintra measures $\sirmse$ within $\mathcal{H}$, or human accuracy independent of the environment; \siinter measures $\sirmse$ between pixels in $\mathcal{H}$ and in $\mathcal{E}$, or human accuracy w.r.t. the environment. We include derivations in the supplemental material.

\begin{table*}[t]
\centering
{\small
\begin{tabular}{lllrrrrrrr}
\toprule
Methods & Dataset & two-view? & \sifull & \sienv & \sihuman  & \siintra  & \siinter & \textbf{RMSE} & \textbf{Rel}
 \\
\midrule
Russell~\etal~\cite{russell2014video} & - & Yes & 2.146 & 2.021 & 2.207 & 2.206 & 2.093 & 2.520 & 0.772 \\
DeMoN~\cite{ummenhofer2017demon} & RGBD+MVS & Yes & 0.338 & 0.302 & 0.360 & 0.293 & 0.384 & 0.866 & 0.220 \\
Chen~\etal~\cite{chen2016single} & NYU+DIW & No & 0.441 & 0.398 & 0.458 & 0.408 & 0.470 & 1.004 & 0.262  \\
Laina~\etal~\cite{laina2016deeper} & NYU & No & 0.358 & 0.356 & 0.349 & 0.270 & 0.377 & 0.947 & 0.223\\
Xu~\etal~\cite{xu2018monocular} & NYU & No & 0.427 & 0.419 & 0.411 & 0.302 & 0.451 & 1.085 & 0.274\\
Fu~\etal~\cite{fu2018deep} & NYU & No & 0.351 & 0.357 & 0.334 & 0.257 & 0.360 & 0.925 & 0.194 \\
\midrule
\rowcolor[gray]{0.95} $I$ & \DatasetShort & No & 0.318 & 0.334 & 0.294 & 0.227 & 0.319 & 0.840 & 0.204 \\
\rowcolor[gray]{0.95} $IFCM$ & \DatasetShort & Yes & 0.316 & 0.330 & 0.302 & 0.228 & 0.323 & 0.843 & 0.206 \\
\rowcolor[gray]{0.95} $I \DPP  M$  & \DatasetShort & Yes & 0.246 & 0.225 & 0.260 & 0.233 & 0.273 & 0.635 & 0.136 \\
\rowcolor[gray]{0.95}$I \DPP  C M$ (w/o  d. cleaning)  & \DatasetShort & Yes  & 0.272 & 0.238 & 0.293 & 0.258 & 0.282 & 0.688 & 0.147 \\
\rowcolor[gray]{0.95} $I \DPP  C M$ & \DatasetShort & Yes & 0.232 & 0.203 & 0.252 & 0.224 & 0.262 & 0.570 & 0.129 \\
\rowcolor[gray]{0.95} $I \DPP  C M K$& \DatasetShort & Yes & \textbf{0.221} & \textbf{0.195} & \textbf{0.238} & \textbf{0.215} & \textbf{0.247} & \textbf{0.541} & \textbf{0.125} \\
\bottomrule
\end{tabular}
}
\vspace{0.2em}
\caption{{\bf Results on TUM RGBD datasets.}  Different \sirmse metrics as well as standard RMSE and relative error (Rel) are reported. We evaluate our models  (light gray background) under different input configurations, as described in Table~\ref{tb:mc_test}. \emph{w/o d. cleaning} indicates the model is trained using raw MVS depth predictions as supervision, without our depth cleaning method. Dataset `-' indicates the method is not learning based. Lower is better for all error  metrics.} \label{tb:tum_rgbd_test}
\vspace{-0.75em}
\end{table*}

\begin{figure*}[t!]
 \centering
\centering
	\vspace*{0.1em} 
    \begin{subfigure}[b]{0.16\textwidth}
        \includegraphics[width=\textwidth]{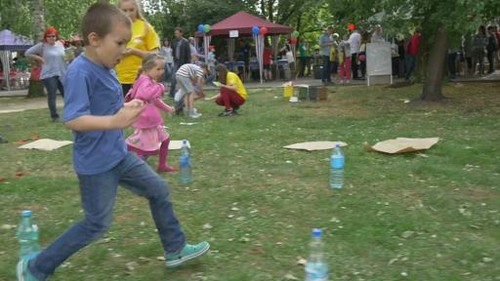}
   \end{subfigure} \hspace*{-0.8em}
    ~
    \begin{subfigure}[b]{0.16\textwidth}
        \includegraphics[width=\textwidth]{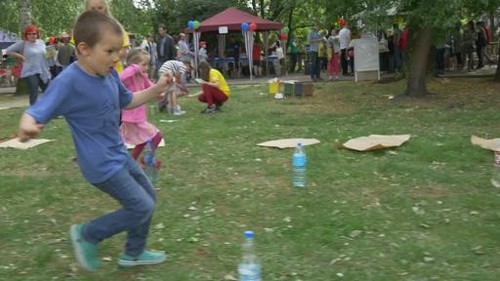}
    \end{subfigure} \hspace*{-0.8em}
    ~
    \begin{subfigure}[b]{0.16\textwidth}
       \includegraphics[width=\textwidth]{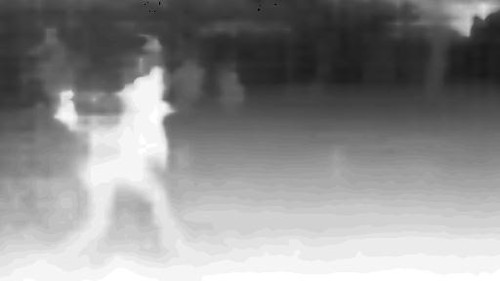}
    \end{subfigure} \hspace*{-0.8em}
    ~
    \begin{subfigure}[b]{0.16\textwidth}
        \includegraphics[width=\textwidth]{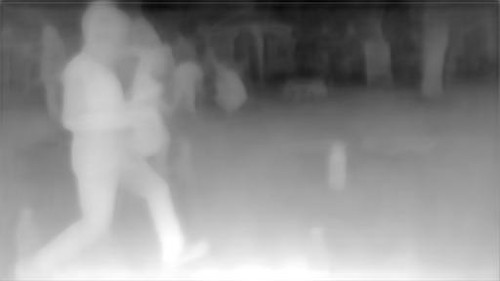}
    \end{subfigure} \hspace*{-0.8em}
    ~    
    \begin{subfigure}[b]{0.16\textwidth}
        \includegraphics[width=\textwidth]{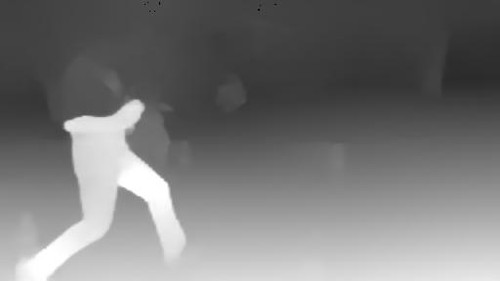}
    \end{subfigure}  \hspace*{-0.8em}
    ~
    \begin{subfigure}[b]{0.16\textwidth}
        \includegraphics[width=\textwidth]{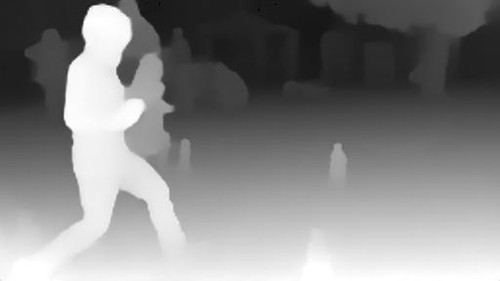}
   \end{subfigure} 
    \begin{subfigure}[b]{0.16\textwidth}
        \includegraphics[width=\textwidth]{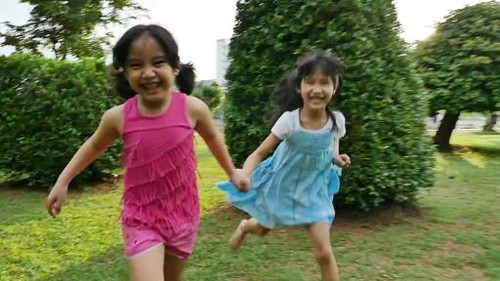}
   \end{subfigure} \hspace*{-0.8em}
    ~
    \begin{subfigure}[b]{0.16\textwidth}
        \includegraphics[width=\textwidth]{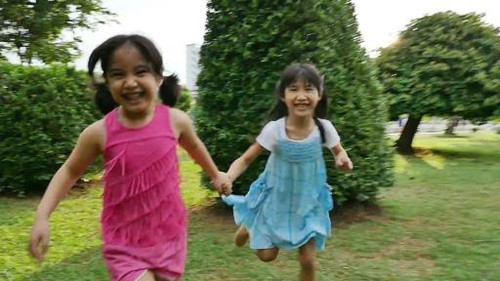}
    \end{subfigure} \hspace*{-0.8em}
    ~
    \begin{subfigure}[b]{0.16\textwidth}
       \includegraphics[width=\textwidth]{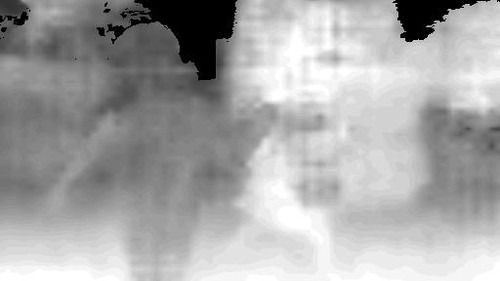}
    \end{subfigure} \hspace*{-0.8em}
    ~
    \begin{subfigure}[b]{0.16\textwidth}
        \includegraphics[width=\textwidth]{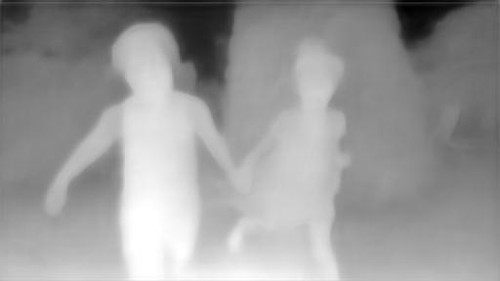}
    \end{subfigure} \hspace*{-0.8em}
    ~    
    \begin{subfigure}[b]{0.16\textwidth}
        \includegraphics[width=\textwidth]{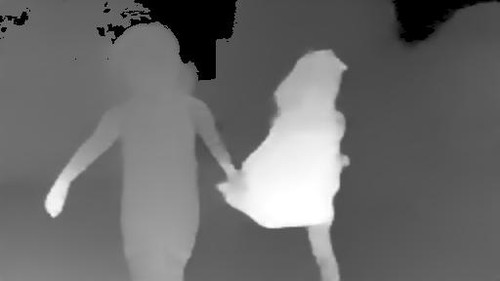}
    \end{subfigure}  \hspace*{-0.8em}
    ~
    \begin{subfigure}[b]{0.16\textwidth}
        \includegraphics[width=\textwidth]{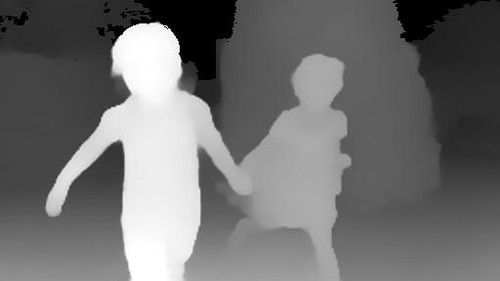}
   \end{subfigure} 
    \begin{subfigure}[b]{0.16\textwidth}
        \includegraphics[width=\textwidth]{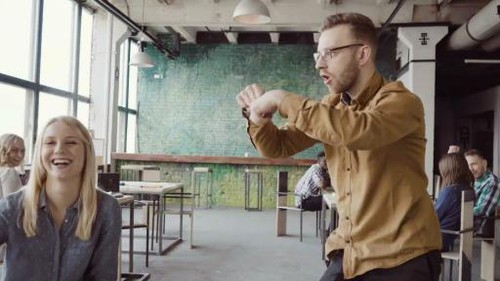}
   \end{subfigure} \hspace*{-0.8em}
    ~
    \begin{subfigure}[b]{0.16\textwidth}
        \includegraphics[width=\textwidth]{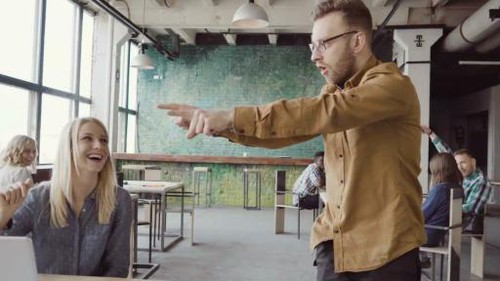}
    \end{subfigure} \hspace*{-0.8em}
    ~
    \begin{subfigure}[b]{0.16\textwidth}
       \includegraphics[width=\textwidth]{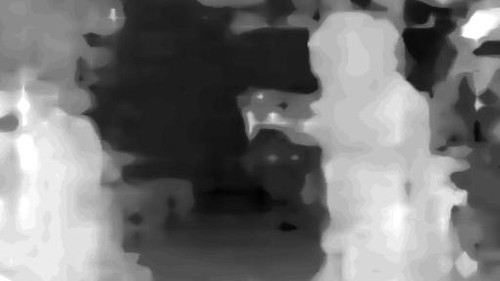}
    \end{subfigure} \hspace*{-0.8em}
    ~
    \begin{subfigure}[b]{0.16\textwidth}
        \includegraphics[width=\textwidth]{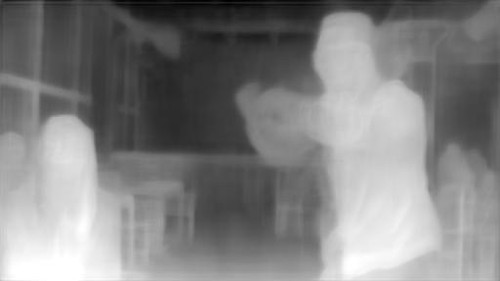}
    \end{subfigure} \hspace*{-0.8em}
    ~    
    \begin{subfigure}[b]{0.16\textwidth}
        \includegraphics[width=\textwidth]{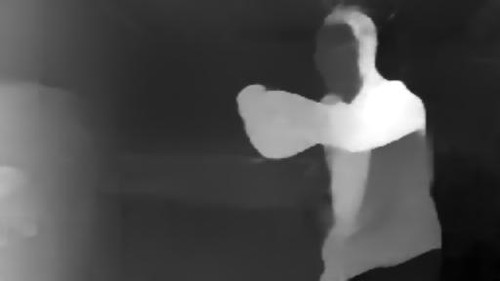}
    \end{subfigure}  \hspace*{-0.8em}
    ~
    \begin{subfigure}[b]{0.16\textwidth}
        \includegraphics[width=\textwidth]{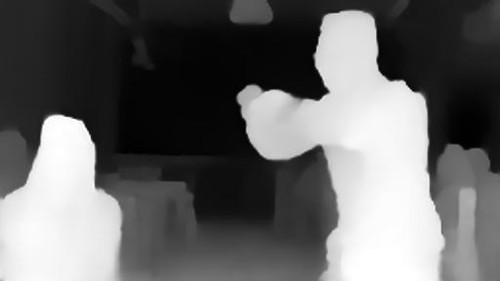}
   \end{subfigure} 
    \begin{subfigure}[b]{0.16\textwidth}
        \includegraphics[width=\textwidth]{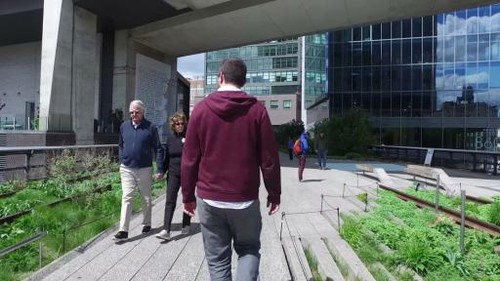}
   \end{subfigure} \hspace*{-0.8em}
    ~
    \begin{subfigure}[b]{0.16\textwidth}
        \includegraphics[width=\textwidth]{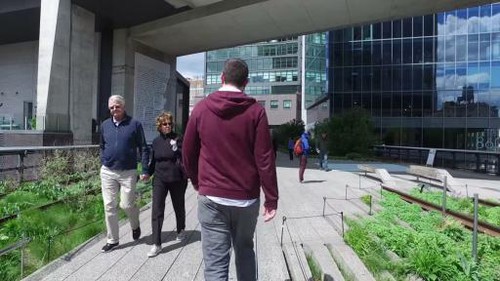}
    \end{subfigure} \hspace*{-0.8em}
    ~
    \begin{subfigure}[b]{0.16\textwidth}
       \includegraphics[width=\textwidth]{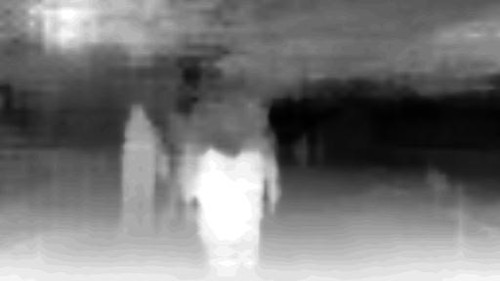}
    \end{subfigure} \hspace*{-0.8em}
    ~
    \begin{subfigure}[b]{0.16\textwidth}
        \includegraphics[width=\textwidth]{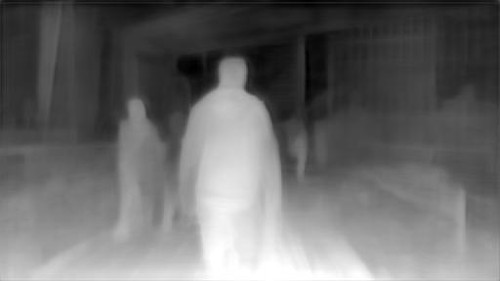}
    \end{subfigure} \hspace*{-0.8em}
    ~    
    \begin{subfigure}[b]{0.16\textwidth}
        \includegraphics[width=\textwidth]{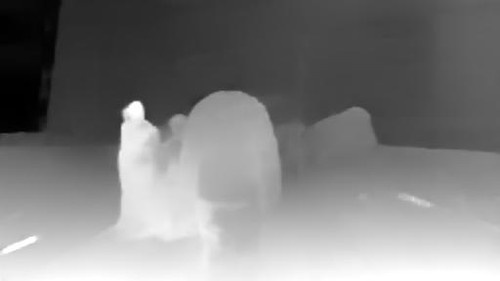}
    \end{subfigure}  \hspace*{-0.8em}
    ~
    \begin{subfigure}[b]{0.16\textwidth}
        \includegraphics[width=\textwidth]{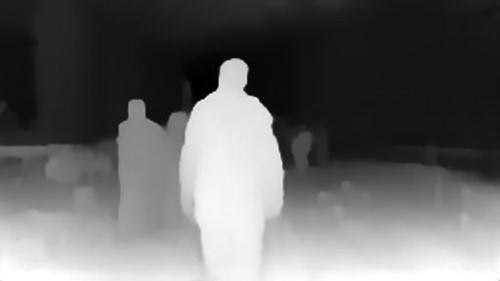}
   \end{subfigure} 
    \begin{subfigure}[b]{0.16\textwidth}
        \includegraphics[width=\textwidth]{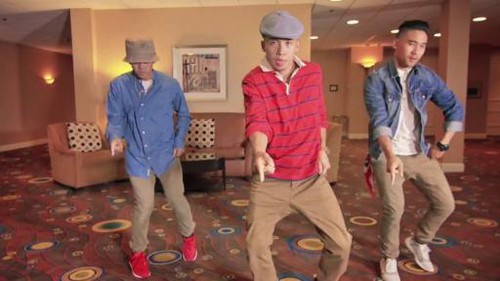}
         \caption{$\Iref$}  \vspace{-0.4em}
   \end{subfigure} \hspace*{-0.8em}
    ~
    \begin{subfigure}[b]{0.16\textwidth}
        \includegraphics[width=\textwidth]{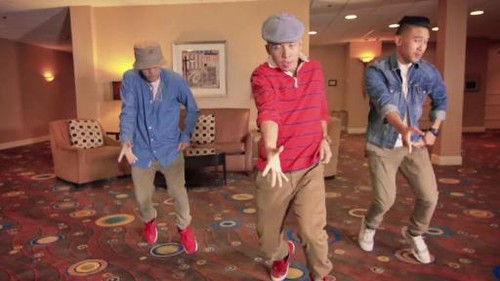}
         \caption{$\Isrc$}  \vspace{-0.4em}
    \end{subfigure} \hspace*{-0.8em}
    ~
    \begin{subfigure}[b]{0.16\textwidth}
       \includegraphics[width=\textwidth]{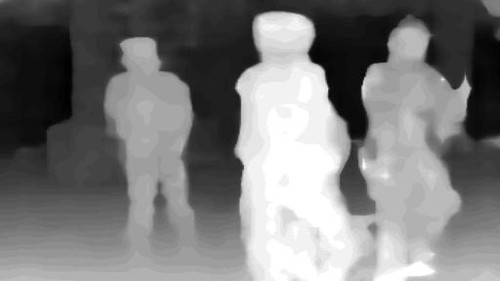}
         \caption{DORN~\cite{fu2018deep}}  \vspace{-0.4em}
    \end{subfigure} \hspace*{-0.8em}
    ~
    \begin{subfigure}[b]{0.16\textwidth}
        \includegraphics[width=\textwidth]{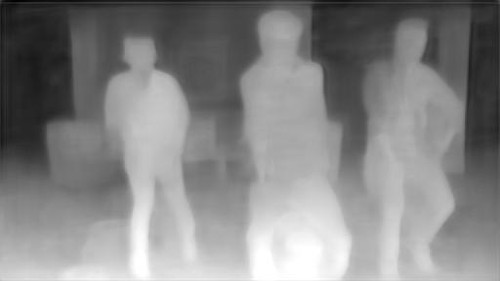}
         \caption{Chen~\etal~\cite{chen2016single}} \vspace{-0.4em}
    \end{subfigure} \hspace*{-0.8em}
    ~    
    \begin{subfigure}[b]{0.16\textwidth}
        \includegraphics[width=\textwidth]{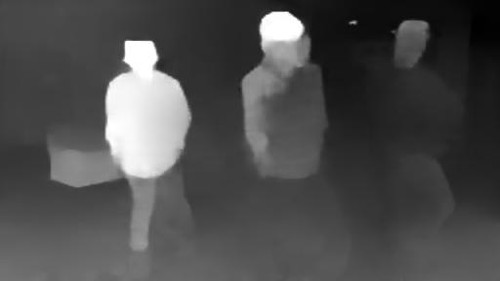}
         \caption{DeMoN~\cite{ummenhofer2017demon}} \vspace{-0.4em}
    \end{subfigure}  \hspace*{-0.8em}
    ~
    \begin{subfigure}[b]{0.16\textwidth}
        \includegraphics[width=\textwidth]{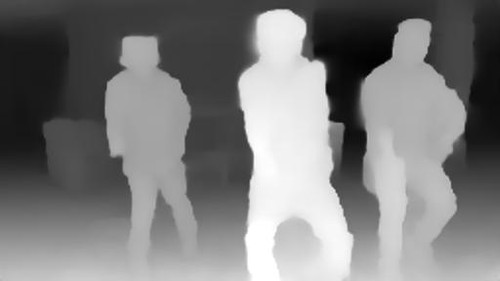}
         \caption{Ours (full)} 
         \vspace{-0.4em}
   \end{subfigure} 
  	\caption{\textbf{Comparisons on Internet video clips with moving cameras and people.} From left to right: (a) reference image, (b) source image, (c) DORN~\cite{fu2018deep}, (d) Chen~\etal~\cite{chen2016single}, (e) DeMoN~\cite{ummenhofer2017demon}, (f) our full method. } \label{fig:demo_compare}
        \vspace{-0.35em}%
\end{figure*}

\subsection{Evaluation on the \DatasetShort test set}
We evaluated our method on our \DatasetShort test set, which consists of more than 29K images taken from 756  video clips. Processed MVS depth values $\DMVS$ obtained by our pipeline (see Sec.~\ref{sec:dataset}) are considered as ground truth.

To quantify the importance of our designed model's input, we compare the performance of several models, each trained on our \DatasetShort dataset with a different input configuration. The two main configurations are: (i) a single-view model (input is RGB image) and (ii) our full two-frame model, where the input includes a reference image, an initial masked depth map $\DPP$, a confidence map $C$, and a human mask $M$. We also perform ablation studies by replacing the input depth with optical flow $F$, removing $C$ from the input, and adding a human keypoint map $K$. %

Quantitative evaluations are shown in Table~\ref{tb:mc_test}. By comparing rows (\Romannum{1}), (\Romannum{3}) and (\Romannum{4}), it is clear that adding the initial depth of environment as well as a confidence map significantly improves the performance for both human and non-human regions. Adding human keypoint locations to the network input further improves performance.
Note that if we input an optical flow field to the network instead of depth (\Romannum{2}), the performance is only on  par with the single view method. The mapping from 2D optical flow to depth depends on the relative camera poses, which are not given to the network. This result indicates that the network is not able to implicitly learn the relative poses and extract the depth information.

Fig.~\ref{fig:prediction_mc_test} shows qualitative comparisons between our single-view model ($I$) and our full model ($I\DPP CMK$). Our full model results are more accurate in both human regions (e.g., first column) and non-human regions (e.g., second column). In addition, the depth relations between people and their surroundings are improved in all examples.

\subsection{Evaluation on TUM RGBD dataset}
We used a subset of the TUM RGBD dataset~\cite{sturm2012benchmark}, which contains indoor scenes of people performing complex actions, captured from different camera poses. Sample images from this dataset are shown in Fig.~\ref{fig:tum_visual_compare}(a-b).

To run our model, we first estimate camera poses using ORB-SLAM2~\footnote{We found estimates from ORB-SLAM2 to be better synchronized with the RGB images than the ground truth poses provided by the TUM dataset.}. In some cases, due to severe low image quality, motion blur and rolling shutter effects, the estimated camera poses may be incorrect. We manually filter such failures  by inspecting the camera trajectory and point cloud. In total, we obtain 11 valid image sequences with 1,815 images in total for evaluations. %

 We compare our depth predictions (using our \DatasetShort trained models) with several state-of-the-art monocular depth prediction methods trained on indoor NYUv2~\cite{laina2016deeper, xu2018monocular, fu2018deep} and Depth in the Wild (DIW) datasets~\cite{chen2016single}, and the recent two-frame stereo model DeMoN~\cite{ummenhofer2017demon}, which assumes a static scene. We also compare with Video-Popup~\cite{russell2014video}, which deals with dynamic scenes. We use the same image pairs for computing $\DPP$ as inputs to DeMoN and Video-Popup.

\begin{figure}[t]
  \centering
    \begin{tabular}{@{\hspace{-0.0em}}c@{\hspace{-0.0em}}c@{\hspace{-0.0em}}}
        \includegraphics[width=0.5\columnwidth]{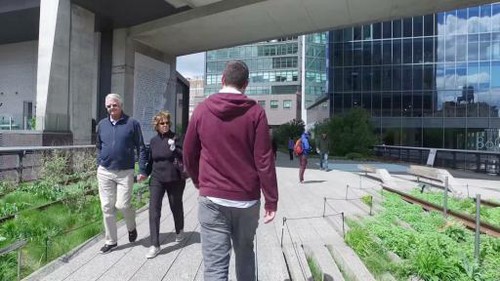} \vspace{-0.1em} & 
        \includegraphics[width=0.5\columnwidth]{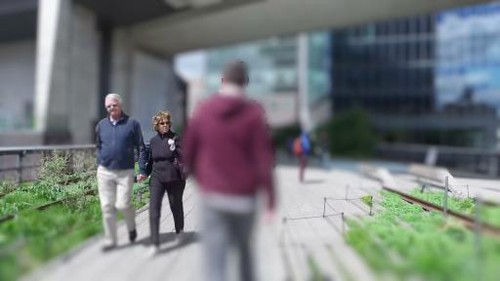}  \vspace{-0.1em} \\
        {\small \text{(a) Input}} \vspace{-0.03em} & 
        {\small \text{(b) Defocus}} \vspace{-0.03em} \\
        \includegraphics[width=0.5\columnwidth]{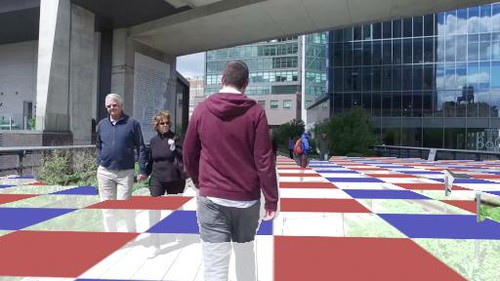} \vspace{-0.1em} & 
        \includegraphics[width=0.5\columnwidth]{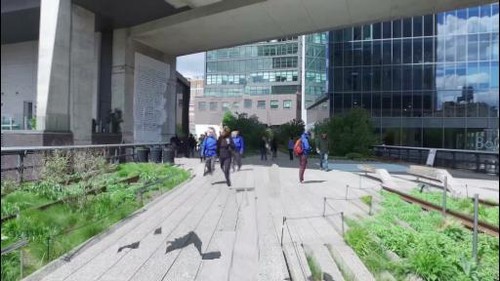} \vspace{-0.1em} \\
        {\small \text{(c) Object insertion}} \vspace{-0.02em} & 
        {\small \text{(d) People removal}} \vspace{-0.02em} \\
        \includegraphics[width=0.5\columnwidth]{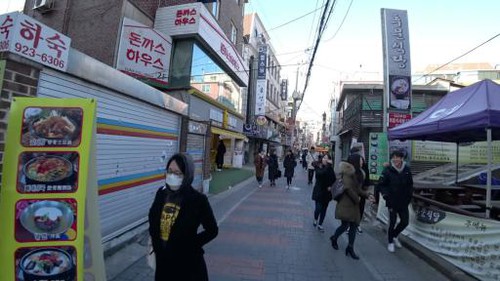}  \vspace{-0.1em} &
        \includegraphics[width=0.5\columnwidth]{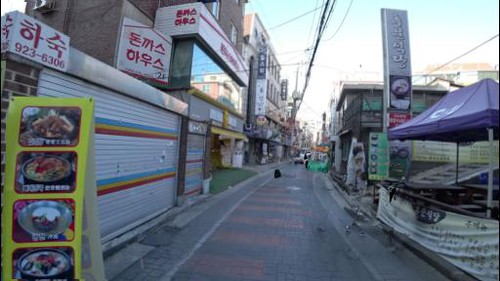} \vspace{-0.1em} \\
        {\small \text{(e) Input}} \vspace{-0.01em} & 
        {\small \text{(f) People removal}} \vspace{-0.01em} \\
    \end{tabular}
  	\caption{\textbf{Depth-based visual effects.} We use our predicted depth maps to apply depth-aware visual effects on (a, e) input images; we show (b) defocus, (c) object insertion, and (d, f) people removal with inpainting results. }\label{fig:depth_effects}
    \vspace{-0.15in}%
\end{figure}

Quantitative comparisons are show in Table~\ref{tb:tum_rgbd_test}, where we report 5 different scale-invariance error measures as well as standard RMSE and relative error; the last two are computed by applying a single scaling factor that aligns the predicted and ground-truth depth in the least-squares sense. Our single-view model already outperforms the other single-view models,%
demonstrating the benefit of the \DatasetShort dataset for training. Note that VideoPopup~\cite{russell2014video} failed to produce meaningful results due to the challenging camera and object motion. Our full model, by making use of the initial (masked) depth map, significantly improves performance for all the error measures. Consistent with our \DatasetShort test set results,  when we use optical flow as input (instead of initial depth map) the performance is only slightly better than the single-view network. Finally, we show the importance of our proposed  ``depth cleaning'' method, applied to the training data (see Eq.~\ref{eq:clean}). Compared to the same model, only trained using the raw MVS depth predictions  as supervision (``w/o d. cleaning''), we see a drop of about 15\% in performance.

Fig.~\ref{fig:tum_visual_compare} shows qualitative comparison between the different methods.  %
Our models' depth predictions (Fig.~\ref{fig:tum_visual_compare}(f-g)) strongly resemble the ground truth and show high level of details and sharp depth discontinuities. This result is a notable improvement over competing methods, which often produce significant errors in both human regions (e.g., legs in the second row of Fig.~\ref{fig:tum_visual_compare}), and non-human regions (e.g., table and ceiling in the last two rows). %

\subsection{Internet videos of dynamic scenes}

We tested our method on challenging Internet videos (downloaded from YouTube and Shutterstock), involving simultaneous natural camera motion and human motion. Our SLAM/SfM pipeline was used to generate sequences ranging from 5 seconds to 15 seconds with smooth and accurate camera trajectories,  after which we apply our method to obtain the required network input buffers.  

We qualitatively compare our full model ($I \DPP  C M K$) with several recent learning based depth prediction models: DORN~\cite{fu2018deep}, Chen~\etal~\cite{chen2016single}, and  DeMoN~\cite{ummenhofer2017demon}. For fair comparisons, we use DORN with a model trained on NYUv2 for indoor videos and a model trained on KITTI for outdoor videos; For \cite{chen2016single}, we use the models trained on both NYUv2 and DIW. For all of our predictions, we use a single model trained from scratch on our \DatasetShort dataset. 

As illustrated in Fig.~\ref{fig:demo_compare}, our depth predictions are significantly better than the baseline methods. In particular, DORN~\cite{fu2018deep} has very limited generalization to Internet videos, and Chen~\etal~\cite{chen2016single}, which is mainly trained on Internet photos, is not able to capture accurate depth. DeMoN often produces incorrect depth, especially in human regions, as it designed for static scenes. Our predicted depth maps depict accurate depth ordering both between people and other objects in the scene (e.g., between people and buildings, fourth row of Fig.~\ref{fig:demo_compare}), and within human regions (such as the arms and legs of people in the first three rows of Fig.~\ref{fig:demo_compare}).

\begin{figure}[t]
  \centering
    \begin{tabular}{@{\hspace{-0.2em}}c@{\hspace{-0.2em}}c@{\hspace{-0.2em}}c@{\hspace{-0.2em}}}
        \includegraphics[width=0.32\columnwidth]{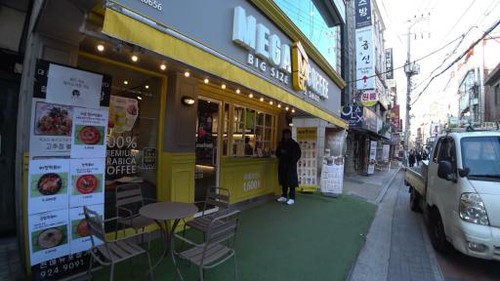} \vspace{-0.05em} & 
        \includegraphics[width=0.32\columnwidth]{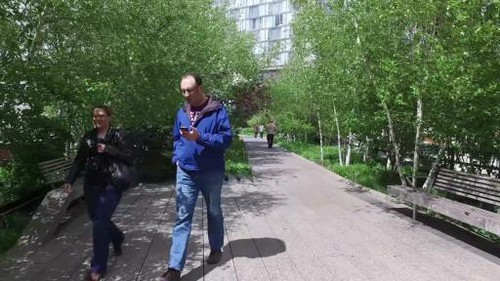}  \vspace{-0.05em} &
        \includegraphics[width=0.32\columnwidth]{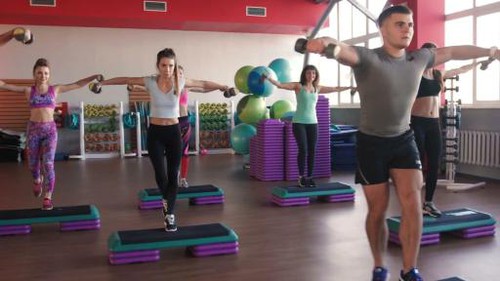} \vspace{-0.05em} \\
        \includegraphics[width=0.32\columnwidth]{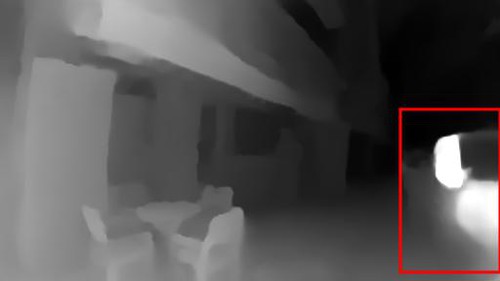} \vspace{-0.05em} & 
        \includegraphics[width=0.32\columnwidth]{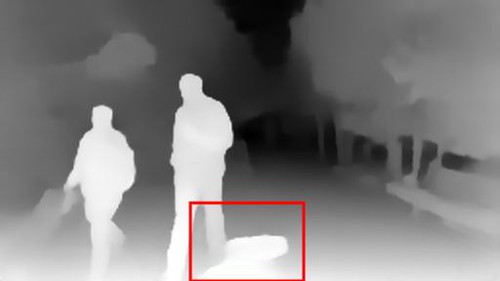}  \vspace{-0.05em} &
        \includegraphics[width=0.32\columnwidth]{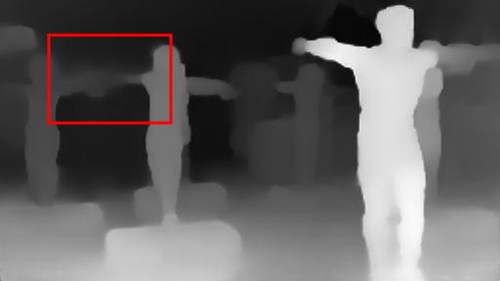} \vspace{-0.05em} 
    \end{tabular}
    \caption{\textbf{Failure cases.} Moving, non-human objects such as cars and shadows can cause bad estimates (left and middle, boxed);  fine structures such as limbs may be blurred for distant people in challenging poses (right, boxed).}
    \label{fig:failures}
    \vspace{-0.8em}
\end{figure}

\medskip
\noindent \textbf{Depth-based visual effects.}
Our depth can be used to apply a range of depth-based visual effects. Fig.~\ref{fig:depth_effects} shows depth-based defocus, insertion of synthetic 3D graphics, and removal of nearby humans with inpainting. See the supplemental material for additional examples, including mono-to-stereo conversion. 

The depth estimates are sufficiently stable over time to allow inpainting from frames elsewhere in the video. To use a frame for inpainting, we construct a triangle heightfield from the depth map, texture the heightfield with the video frame, and render the heightfield from the target frame using the relative camera transformation. 
Fig.~\ref{fig:depth_effects} (d, f) show the results of inpainting two street scenes. Humans near the camera are removed using the human mask $M$, and holes are filled with colors from up to 200 frames later in the video. Some artifacts are visible in areas the human mask misses, such as shadows on the ground.  

\section{Discussion and Conclusion}

We  demonstrated the power of a learning-based approach for predicting dense depth of dynamic scenes where a monocular camera and people are freely moving. We make a new source of data available for training: a large corpus of Mannequin Challenge videos from YouTube, in which the camera moves around and people ``frozen'' in natural poses. We showed how to obtain reliable depth supervision from such noisy data, and demonstrated that our models significantly improve over state-of-the-art methods. %

Our approach still has limitations. We assume known camera poses, which may difficult to infer if moving objects cover most of the scene. In addition, the predicted depth may be inaccurate for non-human, moving regions such as cars and shadows (Fig.~\ref{fig:failures}). Our approach also only uses two views, sometimes leading to temporally inconsistent depth estimates. However,  we hope this work can guide and trigger further progress in monocular dense reconstruction of dynamic scenes.

\newpage
{\small
\bibliographystyle{ieee}
\bibliography{refs}

\begin{thebibliography}{10}\itemsep=-1pt

\bibitem{Bogo2016KeepIS}
F.~Bogo, A.~Kanazawa, C.~Lassner, P.~V. Gehler, J.~Romero, and M.~J. Black.
\newblock {Keep it SMPL: Automatic Estimation of 3D Human Pose and Shape from a
  Single Image}.
\newblock In {\em Proc. European Conf. on Computer Vision (ECCV)}, 2016.

\bibitem{chang2017matterport3d}
A.~Chang, A.~Dai, T.~Funkhouser, M.~Halber, M.~Niessner, M.~Savva, S.~Song,
  A.~Zeng, and Y.~Zhang.
\newblock {Matterport3D}: Learning from {RGB-D} data in indoor environments.
\newblock {\em Int. Conf. on 3D Vision (3DV)}, 2017.

\bibitem{chen2016single}
W.~Chen, Z.~Fu, D.~Yang, and J.~Deng.
\newblock Single-image depth perception in the wild.
\newblock In {\em Neural Information Processing Systems}, pages 730--738, 2016.

\bibitem{dai2017scannet}
A.~Dai, A.~X. Chang, M.~Savva, M.~Halber, T.~A. Funkhouser, and M.~Niessner.
\newblock {ScanNet}: {R}ichly-annotated {3D} reconstructions of indoor scenes.
\newblock In {\em Proc. Computer Vision and Pattern Recognition (CVPR)}, 2017.

\bibitem{Dou2016Fusion4DRP}
M.~Dou, S.~Khamis, Y.~Degtyarev, P.~L. Davidson, S.~R. Fanello, A.~Kowdle,
  S.~Orts, C.~Rhemann, D.~Kim, J.~Taylor, P.~Kohli, V.~Tankovich, and S.~Izadi.
\newblock {Fusion4D}: real-time performance capture of challenging scenes.
\newblock {\em ACM Trans. Graphics}, 35:114:1--114:13, 2016.

\bibitem{eigen2014depth}
D.~Eigen, C.~Puhrsch, and R.~Fergus.
\newblock Depth map prediction from a single image using a multi-scale deep
  network.
\newblock In {\em Neural Information Processing Systems}, pages 2366--2374,
  2014.

\bibitem{fu2018deep}
H.~Fu, M.~Gong, C.~Wang, K.~Batmanghelich, and D.~Tao.
\newblock Deep ordinal regression network for monocular depth estimation.
\newblock In {\em Proc. Computer Vision and Pattern Recognition (CVPR)}, 2018.

\bibitem{Godard2017UnsupervisedMD}
C.~Godard, O.~M. Aodha, and G.~J. Brostow.
\newblock Unsupervised monocular depth estimation with left-right consistency.
\newblock In {\em Proc. Computer Vision and Pattern Recognition (CVPR)}, 2017.

\bibitem{Gler2018DensePoseDH}
R.~A. G{\"u}ler, N.~Neverova, and I.~Kokkinos.
\newblock {DensePose: Dense Human Pose Estimation In The Wild}.
\newblock {\em Proc. Computer Vision and Pattern Recognition (CVPR)}, 2018.

\bibitem{hartley2003multiple}
R.~Hartley and A.~Zisserman.
\newblock {\em Multiple view geometry in computer vision}.
\newblock Cambridge university press, 2003.

\bibitem{howard2002seeing}
I.~P. Howard.
\newblock {\em {Seeing in depth, Vol. 1: Basic mechanisms.}}
\newblock University of Toronto Press, 2002.

\bibitem{huang2018deepmvs}
P.-H. Huang, K.~Matzen, J.~Kopf, N.~Ahuja, and J.-B. Huang.
\newblock {DeepMVS}: Learning multi-view stereopsis.
\newblock In {\em Proc. Computer Vision and Pattern Recognition (CVPR)}, 2018.

\bibitem{ilg2017flownet}
E.~Ilg, N.~Mayer, T.~Saikia, M.~Keuper, A.~Dosovitskiy, and T.~Brox.
\newblock {FlowNet 2.0: Evolution of Optical Flow Estimation With Deep
  Networks}.
\newblock In {\em Proc. Computer Vision and Pattern Recognition (CVPR)}, 2017.

\bibitem{Innmann2016VolumeDeformRV}
M.~Innmann, M.~Zollh{\"o}fer, M.~Niessner, C.~Theobalt, and M.~Stamminger.
\newblock {VolumeDeform}: {R}eal-time volumetric non-rigid reconstruction.
\newblock In {\em Proc. European Conf. on Computer Vision (ECCV)}, 2016.

\bibitem{irani1996parallax}
M.~Irani and P.~Anandan.
\newblock Parallax geometry of pairs of points for 3d scene analysis.
\newblock In {\em Proc. European Conf. on Computer Vision (ECCV)}, pages
  17--30, 1996.

\bibitem{kanazawa2018end}
A.~Kanazawa, M.~J. Black, D.~W. Jacobs, and J.~Malik.
\newblock End-to-end recovery of human shape and pose.
\newblock In {\em Proc. Computer Vision and Pattern Recognition (CVPR)}, 2018.

\bibitem{laina2016deeper}
I.~Laina, C.~Rupprecht, V.~Belagiannis, F.~Tombari, and N.~Navab.
\newblock Deeper depth prediction with fully convolutional residual networks.
\newblock In {\em Int. Conf. on 3D Vision (3DV)}, pages 239--248, 2016.

\bibitem{Lassner2017UniteTP}
C.~Lassner, J.~Romero, M.~Kiefel, F.~Bogo, M.~J. Black, and P.~V. Gehler.
\newblock Unite the people: {C}losing the loop between {3D} and {2D} human
  representations.
\newblock In {\em Proc. Computer Vision and Pattern Recognition (CVPR)}, 2017.

\bibitem{li2018megadepth}
Z.~Li and N.~Snavely.
\newblock {MegaDepth: Learning Single-View Depth Prediction from Internet
  Photos}.
\newblock In {\em Proc. Computer Vision and Pattern Recognition (CVPR)}, 2018.

\bibitem{lv2018learning}
Z.~Lv, K.~Kim, A.~Troccoli, D.~Sun, J.~M. Rehg, and J.~Kautz.
\newblock Learning rigidity in dynamic scenes with a moving camera for 3d
  motion field estimation.
\newblock {\em Proc. European Conf. on Computer Vision (ECCV)}, 2018.

\bibitem{mahjourian2018unsupervised}
R.~Mahjourian, M.~Wicke, and A.~Angelova.
\newblock {Unsupervised Learning of Depth and Ego-Motion from Monocular Video
  Using 3D Geometric Constraints}.
\newblock In {\em Proc. Computer Vision and Pattern Recognition (CVPR)}, 2018.

\bibitem{mees16iros}
O.~Mees, A.~Eitel, and W.~Burgard.
\newblock {Choosing Smartly: Adaptive Multimodal Fusion for Object Detection in
  Changing Environments}.
\newblock In {\em Int. Conf. on Intelligent Robots and Systems (IROS)}, 2016.

\bibitem{Mehta2017VNectR3}
D.~Mehta, S.~Sridhar, O.~Sotnychenko, H.~Rhodin, M.~Shafiei, H.-P. Seidel,
  W.~Xu, D.~Casas, and C.~Theobalt.
\newblock {VNect: Real-time 3D Human Pose Estimation with a Single RGB Camera}.
\newblock {\em ACM Trans. Graphics}, 36:44:1--44:14, 2017.

\bibitem{mur2017orb}
R.~Mur-Artal and J.~D. Tard{\'o}s.
\newblock {Orb-Slam2}: An open-source slam system for monocular, stereo, and
  {RGB-D} cameras.
\newblock {\em IEEE Transactions on Robotics}, 33(5):1255--1262, 2017.

\bibitem{newcombe2015dynamicfusion}
R.~A. Newcombe, D.~Fox, and S.~M. Seitz.
\newblock {DynamicFusion: Reconstruction and tracking of non-rigid scenes in
  real-time}.
\newblock In {\em Proc. Computer Vision and Pattern Recognition (CVPR)}, 2015.

\bibitem{ni2011rgbd}
B.~Ni, G.~Wang, and P.~Moulin.
\newblock {RGBD-HuDaAct: A color-depth video database for human daily activity
  recognition}.
\newblock In {\em Proc. ICCV Workshops}, 2011.

\bibitem{Park20103DRO}
H.~S. Park, T.~Shiratori, I.~A. Matthews, and Y.~Sheikh.
\newblock {3D Reconstruction of a Moving Point from a Series of 2D
  Projections}.
\newblock In {\em Proc. European Conf. on Computer Vision (ECCV)}, 2010.

\bibitem{Pavlakos2017CoarsetoFineVP}
G.~Pavlakos, X.~Zhou, K.~G. Derpanis, and K.~Daniilidis.
\newblock Coarse-to-fine volumetric prediction for single-image 3{D} human
  pose.
\newblock {\em Proc. Computer Vision and Pattern Recognition (CVPR)}, 2017.

\bibitem{ranftl2016dense}
R.~Ranftl, V.~Vineet, Q.~Chen, and V.~Koltun.
\newblock Dense monocular depth estimation in complex dynamic scenes.
\newblock In {\em Proc. Computer Vision and Pattern Recognition (CVPR)}, 2016.

\bibitem{rematas2018soccer}
K.~Rematas, I.~Kemelmacher-Shlizerman, B.~Curless, and S.~Seitz.
\newblock Soccer on your tabletop.
\newblock In {\em Proc. Computer Vision and Pattern Recognition (CVPR)}, June
  2018.

\bibitem{russell2014video}
C.~Russell, R.~Yu, and L.~Agapito.
\newblock Video pop-up: Monocular 3d reconstruction of dynamic scenes.
\newblock In {\em Proc. European Conf. on Computer Vision (ECCV)}, pages
  583--598, 2014.

\bibitem{schonberger2016structure}
J.~L. Schonberger and J.-M. Frahm.
\newblock Structure-from-motion revisited.
\newblock In {\em Proc. Computer Vision and Pattern Recognition (CVPR)}, 2016.

\bibitem{schonberger2016pixelwise}
J.~L. Sch{\"o}nberger, E.~Zheng, J.-M. Frahm, and M.~Pollefeys.
\newblock Pixelwise view selection for unstructured multi-view stereo.
\newblock In {\em Proc. European Conf. on Computer Vision (ECCV)}, pages
  501--518, 2016.

\bibitem{shrivastava2017learning}
A.~Shrivastava, T.~Pfister, O.~Tuzel, J.~Susskind, W.~Wang, and R.~Webb.
\newblock {Learning from Simulated and Unsupervised Images through Adversarial
  Training}.
\newblock In {\em Proc. Computer Vision and Pattern Recognition (CVPR)}, 2017.

\bibitem{silberman2012indoor}
N.~Silberman, D.~Hoiem, P.~Kohli, and R.~Fergus.
\newblock Indoor segmentation and support inference from rgbd images.
\newblock In {\em Proc. European Conf. on Computer Vision (ECCV)}, 2012.

\bibitem{Simon2017KroneckerMarkovPF}
T.~Simon, J.~Valmadre, I.~A. Matthews, and Y.~Sheikh.
\newblock {Kronecker-Markov Prior for Dynamic 3D Reconstruction}.
\newblock {\em Trans. Pattern Analysis and Machine Intelligence},
  39:2201--2214, 2017.

\bibitem{song2016ssc}
S.~Song, F.~Yu, A.~Zeng, A.~X. Chang, M.~Savva, and T.~Funkhouser.
\newblock Semantic scene completion from a single depth image.
\newblock {\em Proc. Computer Vision and Pattern Recognition (CVPR)}, 2017.

\bibitem{sturm2012benchmark}
J.~Sturm, N.~Engelhard, F.~Endres, W.~Burgard, and D.~Cremers.
\newblock A benchmark for the evaluation of {RGB-D SLAM} systems.
\newblock In {\em IEEE/RSJ International Conference on Intelligent Robots and
  Systems (IROS)}, pages 573--580, 2012.

\bibitem{ummenhofer2017demon}
B.~Ummenhofer, H.~Zhou, J.~Uhrig, N.~Mayer, E.~Ilg, A.~Dosovitskiy, and
  T.~Brox.
\newblock {DeMoN: Depth and motion network for learning monocular stereo}.
\newblock In {\em Proc. Computer Vision and Pattern Recognition (CVPR)}, 2017.

\bibitem{Vo2016SpatiotemporalBA}
M.~Vo, S.~G. Narasimhan, and Y.~Sheikh.
\newblock {Spatiotemporal Bundle Adjustment for Dynamic 3D Reconstruction}.
\newblock {\em Proc. Computer Vision and Pattern Recognition (CVPR)}, 2016.

\bibitem{Wang_2018_CVPR}
C.~Wang, J.~Miguel~Buenaposada, R.~Zhu, and S.~Lucey.
\newblock Learning depth from monocular videos using direct methods.
\newblock In {\em Proc. Computer Vision and Pattern Recognition (CVPR)}, 2018.

\bibitem{wikimannequinchallenge}
Wikipedia.
\newblock {Mannequin Challenge}.
\newblock \url{https://en.wikipedia.org/wiki/Mannequin_Challenge}, 2018.

\bibitem{wulff2017optical}
J.~Wulff, L.~Sevilla-Lara, and M.~J. Black.
\newblock Optical flow in mostly rigid scenes.
\newblock In {\em Proc. Computer Vision and Pattern Recognition (CVPR)}, 2017.

\bibitem{XianMonocularRD}
K.~Xian, C.~Shen, Z.~Cao, H.~Lu, Y.~Xiao, R.~Li, and Z.~Luo.
\newblock Monocular relative depth perception with web stereo data supervision.
\newblock In {\em Proc. Computer Vision and Pattern Recognition (CVPR)}, 2018.

\bibitem{xiao2013sun3d}
J.~Xiao, A.~Owens, and A.~Torralba.
\newblock {Sun3D}: A database of big spaces reconstructed using sfm and object
  labels.
\newblock In {\em Proc. Int. Conf. on Computer Vision (ICCV)}, pages
  1625--1632, 2013.

\bibitem{xu2018monocular}
D.~Xu, E.~Ricci, W.~Ouyang, X.~Wang, and N.~Sebe.
\newblock Monocular depth estimation using multi-scale continuous crfs as
  sequential deep networks.
\newblock {\em Trans. Pattern Analysis and Machine Intelligence}, 2018.

\bibitem{yao2018mvsnet}
Y.~Yao, Z.~Luo, S.~Li, T.~Fang, and L.~Quan.
\newblock {MVSNet: Depth Inference for Unstructured Multi-view Stereo}.
\newblock {\em Proc. European Conf. on Computer Vision (ECCV)}, 2018.

\bibitem{ye2014real}
M.~Ye and R.~Yang.
\newblock Real-time simultaneous pose and shape estimation for articulated
  objects using a single depth camera.
\newblock In {\em Proc. Computer Vision and Pattern Recognition (CVPR)}, 2014.

\bibitem{Yin2018GeoNetUL}
Z.~Yin and J.~Shi.
\newblock {GeoNet: Unsupervised Learning of Dense Depth, Optical Flow and
  Camera Pose}.
\newblock In {\em Proc. Computer Vision and Pattern Recognition (CVPR)}, 2018.

\bibitem{Zheng2015SparseD3}
E.~Zheng, D.~Ji, E.~Dunn, and J.-M. Frahm.
\newblock {Sparse Dynamic 3D Reconstruction from Unsynchronized Videos}.
\newblock {\em Proc. Int. Conf. on Computer Vision (ICCV)}, pages 4435--4443,
  2015.

\bibitem{zhou2018deeptam}
H.~Zhou, B.~Ummenhofer, and T.~Brox.
\newblock {DeepTAM: Deep Tracking and Mapping}.
\newblock In {\em Proc. European Conf. on Computer Vision (ECCV)}, 2018.

\bibitem{Zhou2017UnsupervisedLO}
T.~Zhou, M.~Brown, N.~Snavely, and D.~G. Lowe.
\newblock Unsupervised learning of depth and ego-motion from video.
\newblock In {\em Proc. Computer Vision and Pattern Recognition (CVPR)}, 2017.

\bibitem{zhou2018stereo}
T.~Zhou, R.~Tucker, J.~Flynn, G.~Fyffe, and N.~Snavely.
\newblock {Stereo Magnification: Learning view synthesis using multiplane
  images}.
\newblock {\em ACM Trans. Graphics (SIGGRAPH)}, 2018.

\bibitem{zhu2014evaluating}
Y.~Zhu, W.~Chen, and G.~Guo.
\newblock Evaluating spatiotemporal interest point features for depth-based
  action recognition.
\newblock {\em Image and Vision Computing}, 32(8):453--464, 2014.

\bibitem{zollhofer2014real}
M.~Zollh{\"o}fer, M.~Niessner, S.~Izadi, C.~Rehmann, C.~Zach, M.~Fisher, C.~Wu,
  A.~Fitzgibbon, C.~Loop, C.~Theobalt, et~al.
\newblock Real-time non-rigid reconstruction using an {RGB-D} camera.
\newblock {\em ACM Trans. Graphics}, 33(4):156, 2014.

\end{thebibliography}
}

\end{document}